\documentclass{article} % For LaTeX2e
\usepackage{iclr2026_conference,times}
\usepackage{tikz}
\usepackage{caption}
\usepackage{booktabs}
\usepackage{graphicx}
\usepackage{enumitem}
\usepackage{amssymb}
\usepackage{wrapfig}
\usepackage{sidecap}
\usepackage{placeins}
\usepackage{algorithm, algpseudocode}
\usetikzlibrary{calc}

\usepackage{amsmath,amsfonts,bm}

\def\eqref#1{equation~\ref{#1}}
\def\1{\bm{1}}

\DeclareMathAlphabet{\mathsfit}{\encodingdefault}{\sfdefault}{m}{sl}
\SetMathAlphabet{\mathsfit}{bold}{\encodingdefault}{\sfdefault}{bx}{n}

\DeclareMathOperator*{\argmax}{arg\,max}

\usepackage[hidelinks]{hyperref}
\usepackage{url}

\title{Partial Soft-Matching Distance for Neural Representational Comparison with Partial Unit Correspondence}

\author{Chaitanya Kapoor \\
Department of Cognitive Science \\
University of California, San Diego \\
La Jolla, CA, 92093 \\
\texttt{chkapoor@ucsd.edu}
\And
Alex H. Williams \\
Center for Neural Science, New York University \\
Center for Computational Neuroscience, Flatiron Institute \\
New York City, NY, 10003 \\
\texttt{alex.h.williams@nyu.edu}
\And
Meenakshi Khosla \\
Department of Cognitive Science \\
Department of Computer Science and Engineering \\
University of California, San Diego \\
La Jolla, CA, 92093 \\
\texttt{mkhosla@ucsd.edu}
}

\iclrfinalcopy % Uncomment for camera-ready version, but NOT for submission.
\begin{document}

\maketitle

\begin{abstract}
Representational similarity metrics typically force all units to be matched, making them susceptible to noise and outliers common in neural representations. We extend the soft-matching distance to a partial optimal transport setting that allows some neurons to remain unmatched, yielding rotation-sensitive but robust correspondences. This partial soft-matching distance provides theoretical advantages---relaxing strict mass conservation while maintaining interpretable transport costs---and practical benefits through efficient neuron ranking in terms of cross-network alignment without costly iterative recomputation. In simulations, it preserves correct matches under outliers and reliably selects the correct model in noise-corrupted identification tasks. On fMRI data, it automatically excludes low-reliability voxels and produces voxel rankings by alignment quality that closely match computationally expensive brute-force approaches. It achieves higher alignment precision across homologous brain areas than standard soft-matching, which is forced to match all units regardless of quality. In deep networks, highly matched units exhibit similar maximally exciting images, while unmatched units show divergent patterns. This ability to partition by match quality enables focused analyses, \emph{e.g.,} testing whether networks have privileged axes even within their most aligned subpopulations. Overall, partial soft-matching provides a principled and practical method for representational comparison under partial correspondence.\blfootnote{All code is publicly available at: \textcolor{magenta}{\href{https://github.com/NeuroML-Lab/partial-metric/}{https://github.com/NeuroML-Lab/partial-metric/}}}
\end{abstract}

% introduction
\section{Introduction}
%Understanding how design choices such as training objectives and architectural decisions influence neural representations requires comparing how different systems encode information. A fundamental challenge in this comparison is determining which computational units correspond across systems: do specific neurons in one network implement similar functions to neurons in another? This question is central to understanding whether different systems converge on similar computational solutions.
Understanding how design choices (\emph{e.g.,} training objectives, architecture) shape neural representations requires comparing how different systems encode information. A fundamental challenge in this comparison is determining which computational units correspond across systems: do specific neurons implement similar functions across networks? This is central to understanding whether different systems converge to similar computational solutions. Most existing representational similarity metrics, such as CKA~\citep{kornblith2019similarity}, RSA~\citep{kriegeskorte2008representational}, Procrustes distance~\citep{williams2021generalized,Ding2021}, and CCA variants~\citep{raghu2017svcca}, are rotation-invariant---they measure geometric similarity while ignoring the specific axes along which information is encoded. This limitation prevents us from understanding neuron-level correspondence and whether systems share axis-aligned representations. 

%\begin{figure}
%    \centering
%    \includegraphics[width=0.75\columnwidth]{./figures/iclr_schematic_left.pdf}
%    \caption{\textbf{TODO.}}
%    \label{fig:placeholder}
%\end{figure}
%\begin{figure}[htbp!]
%  \centering
%  \includegraphics[width=0.43\columnwidth]{./figures/iclr_schematic_left.pdf}\hfill
%  \raisebox{3em}{
%  \includegraphics[width=0.53\columnwidth]{./figures/iclr_schematic_right.pdf}
%  }
%  \caption{\textbf{TODO}}
%  \label{fig:placeholder}
%\end{figure}

\begin{figure}[htbp!]
  \centering
  \begin{tikzpicture}
    % Place the two images side by side
    \node (left) at (0,0) {\includegraphics[width=0.43\columnwidth]{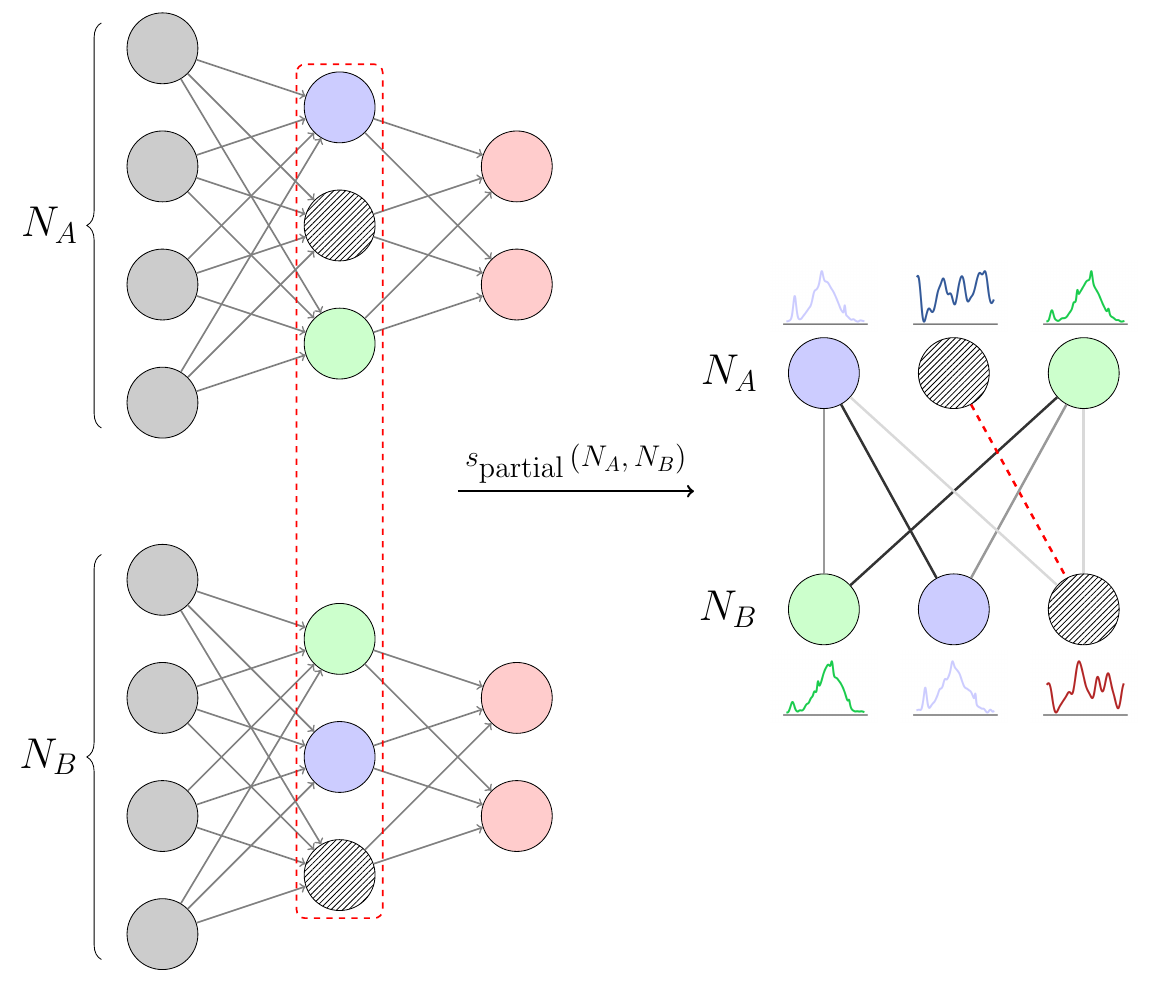}};
    \node (right) at (7,0) {\includegraphics[width=0.53\columnwidth]{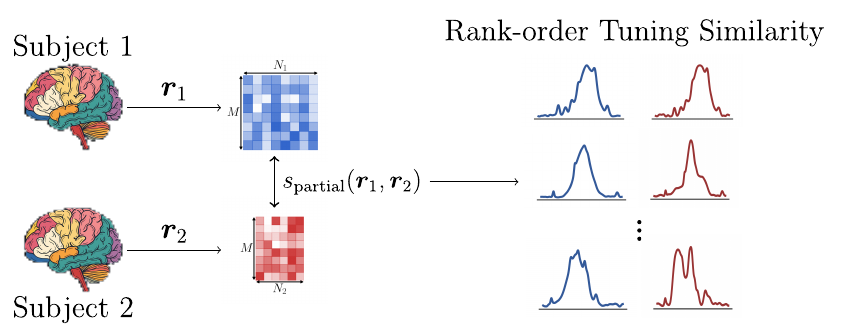}};
    % Labels (A) and (B)
    \node[above] at ($(left.north west) + (1, 0)$) {\large \textbf{A}};
    \node[above] at ($(right.north west) + (0.3, -0.1)$) {\large \textbf{B}};
  \end{tikzpicture}

  \caption{\textbf{Partial Soft-Matching Distance for Matching Tuning Curves.} \textbf{(A)} Two toy networks $N_A$ and $N_B$; the layer of interest for alignment is shown in red. A partial matching recovers one-to-one correspondences between units with highly similar tuning curves. Line color encodes match strength (darker = stronger). By contrast, a purely soft-matching yields a spurious pair (hatched units, red dotted line). \textbf{(B)} The same metric can be used to rank voxel/unit tuning-curve similarity between two subjects’ responses $\{\bm{r}_1, \bm{r}_2\}$, when exposed to the same visual stimulus.}
  \label{fig:schematic}
\end{figure}

To address these challenges, \citet{khosla2024soft} recently proposed a metric based on discrete optimal transport \citep[OT; ][]{peyre2019computational} called the \textit{soft-matching distance} which finds rotation-sensitive correspondences between neurons while remaining invariant to their ordering.
However, this approach requires that all neural units are matched across networks. This requirement may be a crucial limitation in certain settings, since neural populations often contain noisy, inactive, or task-irrelevant units---particularly in biological recordings from fMRI or electrophysiology. Moreover, even task-relevant units may be model-specific, implementing computations unique to a particular architecture or training regime in deep neural networks (DNNs). When comparing networks, we should not expect complete overlap in their functional units. Forcing all units into correspondence may therefore inflate distances and produce misleading alignments.
%However, the soft-matching distance inherits a critical limitation from classical optimal transport: it requires all units to be matched. In practice, neural populations often contain noisy, inactive, or task-irrelevant units---particularly in biological recordings from fMRI or electrophysiology. Additionally, even task-relevant units may be model-specific, implementing computations unique to a particular architecture or training regime. When comparing networks trained on different tasks or with different architectures, we should not expect complete overlap in their functional units. Forcing all units into correspondence inflates distances and produces misleading alignments.

%We introduce the \textbf{unbalanced soft-matching distance}, which extends soft-matching to partial OT (Fig.~\ref{fig:schematic}). This allows a fraction of neurons to remain unmatched while preserving the ability to detect robust correspondences among the remainder. Our method provides several key advantages:
Here, we introduce the \textbf{partial soft-matching distance} (Fig.~\ref{fig:schematic}), which allows a fraction of neurons to remain unmatched while preserving robust correspondences among the remainder. Our method provides several key advantages:
\begin{itemize}[nosep,leftmargin=1.5em]
    \item \textbf{Theoretical robustness:} Relaxing mass conservation allows the metric to handle populations with different numbers of units, where some may lack correspondence (\emph{e.g.}, due to noise).
    \item \textbf{Computational efficiency:} Achieves comparable rankings with a single $\mathcal{O}(n^3 \log n)$ computation, unlike brute-force methods requiring $\mathcal{O}(n^4 \log n)$ operations.
    \item \textbf{Interpretable partitioning:} Separates well-matched from unmatched units, enabling focused analysis of aligned subpopulations.
\end{itemize}

% Our method provides several key advantages:
% \begin{itemize}
%     \item \textbf{Theoretical robustness:} Relaxing mass conservation constraints allows the metric to naturally handle populations with different number of units where not all units have correspondence (\emph{e.g.,} noise, task-irrelevance).  
%     \item \textbf{Computational efficiency:} Unlike brute-force approaches that require $\mathcal{O}(n^{4}\log n)$ operations to rank neuron correspondence, our method achieves comparable results with a single $\mathcal{O}(n^{3}\log n)$ computation.
%     \item \textbf{Interpretable partitioning:} Identify well-matched vs. unmatched units allowing analysis of aligned subpopulations.
% \end{itemize}
%By relaxing mass conservation constraints, the metric naturally handles populations with different numbers of units where not all units have correspondences---whether due to noise, inactivity, task-irrelevance, or network-specific computations that exist in one system but not another.
    %\item \textbf{Computational efficiency:} Unlike brute-force approaches that require $\mathcal{O}(n^{4}\log n)$ operations to rank neuron correspondence, our method achieves comparable results with a single $\mathcal{O}(n^{3}\log n)$ computation.
We demonstrate these advantages through controlled simulations showing correspondence despite spurious neurons, and accurate model identification in noise-corrupted scenarios. In fMRI data from the Natural Scenes Dataset~\citep{allen2022massive}, our method discards low-quality voxels and outperforms standard soft-matching in aligning homologous brain regions across subjects. When applied to DNNs, we find that highly-matched units produce similar maximally exciting images (MEIs) across models, while unmatched units show divergent MEIs, suggesting distinct computational roles. Crucially, filtering unmatched units using partial soft-matching improves alignment over heuristics based on soft-matching correlations, matching the performance of a computationally intensive brute-force method that iteratively removes units in a greedy fashion. This framework provides a principled approach for comparing neural representations under partial correspondence---a common scenario in neuroscience and AI.  
\section{Methods}
The optimal transport (OT) problem finds the minimum-cost mapping between probability distributions, yielding metrics like the soft-matching distance~\citep{khosla2024soft}. However, classical OT requires equal total mass between distributions—a constraint violated in neural recordings where units may be noisy, inactive, or genuinely non-corresponding. We extend the soft-matching distance to handle these realistic scenarios through partial optimal transport.

% \textcolor{red}{\paragraph{Notations.} We define $\bm{p}$ and $\bm{q}$ as empirical probability measures supported on sets of Dirac point masses $\mathcal{X}=\{\bm{x}_i\}_{i=1}^{N_x}$ and $\mathcal{Y}=\{\bm{y}_j\}_{j=1}^{N_y}$ with uniform probability masses. The total \emph{``mass''} of a measure is its $\ell_1$-norm, $||\bm{p}||_1$ and $||\bm{q}||_1$. In standard OT formulations, the linear equality constraints enforce $||\bm{p}||_1 = ||\bm{q}||_1 = 1$. By contrast, the partial OT problem relaxes the equality constraint, and transports only a fraction $s$ of the total mass. Concretely, we solve for the cheapest transport plan that moves a mass $0\leq s\leq \min(||\bm{p}||_1,  ||\bm{q}||_1)$ between $\bm{p}$ and $\bm{q}$.}

\subsection{Soft-matching distance}
Consider two neural populations with $N_x$ and $N_y$ units respectively, each with ``tuning curves'', $\{\bm{x}_i\}_{i=1}^{N_x}$ and $\{\bm{y}_j\}_{j=1}^{N_y}$ taking values in $\mathbb{R}^M$.
Each unit's tuning curve, respectively denoted $\mathbf{x}_i$ and $\mathbf{y}_j$ for the two neural populations, represents a neuron's response over a set of $M$ probe stimuli.
Stacking these tuning curve vectors column-wise produces matrices $\bm{X} \in \mathbb{R}^{M \times N_x}$ and $\bm{Y} \in \mathbb{R}^{M \times N_y}$.

The soft-matching distance treats each population as a uniform empirical measure and quantifies the optimal transport cost between them:
\[
d_T(\bm{X}, \bm{Y}) = \min_{\bm{T} \in \mathcal{T}(N_x,N_y)} \sqrt{\langle \bm{C}, \bm{T} \rangle_F}
\]
where $\bm{C}_{ij} = \|\bm{x}_i - \bm{y}_j\|^2$ is the squared Euclidean transport cost, $\langle\cdot, \cdot\rangle_F$ the Frobenius inner product, and $\mathcal{T}(N_x, N_y)$ is the transportation polytope~\citep{de2013combinatorics}, i.e., the set of all $N_x \times N_y$ nonnegative matrices whose rows each sum to $1/N_x$ and whose columns each sum to $1/N_y$.

This formulation is permutation-invariant yet rotation-sensitive, revealing single-neuron tuning alignment. 
Furthermore, $d_T$ is symmetric and satisfies the triangular inequality, which has been argued to be important for certain analyses of neural representations~\citep{williams2021generalized,pmlr-v221-lange23a}.
The key limitation (which we document in Sections \ref{sec: synthetic} and \ref{sec:applications}) is that the marginal constraints (i.e. that the transport plan $\bm{T}$ lie within the transportation polytope) forces all units to be matched, producing spurious correspondences when populations contain non-corresponding units.
%Consider two neural populations with $N_x$ and $N_y$ units respectively, each with ``tuning curves'' $\{\bm{x}_i, \bm{y}_j\} \in \mathbb{R}^M$ measured over $M$ stimuli. The soft-matching distance treats each population as a uniform empirical measure and solves $
%d_T(\bm{p}, \bm{q}) = \min_{T \in \mathcal{T}(N_x,N_y)} \langle \bm{C}, \bm{T} \rangle_F
%$
%where $\bm{C}_{ij} = \|\bm{x}_i - \bm{y}_j\|^2$ is the squared Euclidean cost, $\langle\cdot, \cdot\rangle_F$ the Frobenius norm, and $\mathcal{T}(N_x, N_y)$ is the transportation polytope~\citep{de2013combinatorics}, i.e., the set of all $N_x \times N_y$ nonnegative matrices whose rows each sum to $1/N_x$ and whose columns each sum to $1/N_y$. 
%
%This formulation is permutation-invariant yet rotation-sensitive, revealing single-neuron tuning alignment. The key limitation is that the marginal constraints force all units to match, producing spurious correspondences when populations contain non-corresponding units.

\subsection{Partial Soft-Matching Distance}
The soft-matching formulation requires the two empirical distributions to have identical total mass and further enforces that \textbf{all} mass must be transported. The partial OT problem extends this by allowing only a pre-specified fraction $0\leq s\leq 1$ of the total mass to be matched at minimal cost.

Formally, for empirical measures with unit total mass, a natural set of admissible couplings is
\begin{equation*}
    \mathcal{T}^{s}(N_x,N_y)
    \;=\;
    \left\{
      \bm{T}\in\mathbb{R}_+^{N_x\times N_y}
      \ \middle|\ 
      \sum_{j=1}^{N_y}\bm{T}_{ij}\le \tfrac{1}{N_x},\quad
      \sum_{i=1}^{N_x}\bm{T}_{ij}\le \tfrac{1}{N_y},\quad
      \sum_{i,j}\bm{T}_{ij}=s
    \right\}.
\end{equation*}
Here, the inequalities on the row/column marginals permit mass to remain unmatched in either population, and the scalar $s$ controls the total matched mass. Since we normalize our populations to have unit total mass, $s$ directly represents the fraction of units that are actually matched. The partial soft-matching distance is then the minimum transport cost over this feasible set,
\begin{equation*}
    d_{\bm{T}}(\bm{X},\bm{YX})
    \;=\;
    \min_{\bm{T}\in\mathcal{T}^{s}(N_x,N_y)}
    \langle\bm{C},\bm{T}\rangle_F,
\end{equation*}
with $\bm{C}$ the usual cost matrix (\emph{e.g.,} $\bm{C}_{ij} = ||\bm{x}_i - \bm{y}_j||^2$). In our formulation, we use pairwise cosine distance as the cost function. Several numerical approaches have been developed to solve partial OT problems~\citep{benamou2015iterative,chizat2018scaling}. More recently, \citet{chapel2020partial} augmented the cost matrix with dummy (or virtual) points which are assigned large transportation cost. All mass routed to these dummy nodes is effectively discarded, which yields an exact partial-matching solution in the augmented formulation.

Partial OT distances do not satisfy the triangle inequality and therefore are not proper metrics.
However, they still provide a symmetric notion of dissimilarity in representation and, as we document in Sections \ref{sec: synthetic} and \ref{sec:applications}, they provide robust and interpretable tool for tuning-level comparisons between neural populations with unequal or noisy measurements.

\subsection{Choosing Optimal Regularization}
\label{sec: optimal-mreg}
To apply partial soft-matching in practice, we must select the hyperparameter $0 \leq s \leq 1$ which determines how much mass to transport between distributions.
This is a key challenge when the abundance of outliers and magnitude of noise in the data are unknown \emph{a priori}. To address this, we adopt an L-curve heuristic~\citep{cultrera2020simple}, inspired by classical regularization methods for ill-posed problems (\emph{e.g.,} Tikhonov regularization). The L-curve captures the tradeoff between transport distance and regularization strength, with the ``elbow'' typically indicating a balanced choice between these competing objectives. Concretely, we define the two-dimensional parametric curve:
%A key challenge lies in ascertaining the amount of mass that must be transported between the distributions, $(\bm{p}, \bm{q})$, when the amount of noise is not known \emph{a priori}. To address this, we draw inspiration from classical regularization techniques from ill-posed inverse problems (\emph{e.g.,} Tikhonov regularization) and select the regularization parameter using an L-curve heuristic~\citep{cultrera2020simple}. The L-curve visualizes the tradeoff between errors (here, distance) and regularization strength. The ``corner'' or ``elbow'' of this graph typically indicates a good balance between these competing objectives. Concretely, define the two-dimensional parametric curve:
\begin{equation*}
    f(s) = (\zeta(s), \rho(s)) \rightarrow  
    \begin{cases}
        \zeta(s) = \langle \bm{T}(s),\ \bm{C}\rangle_F \\
        \rho(s) = 1 - s
    \end{cases}
\end{equation*}
where $\bm{C}$ is the cost matrix and $\bm{T}(s)$ is the optimal transport plan for a match fraction $s\in[0, 1]$. We interpret $\rho(s)$ as the regularization strength---smaller $s$, or conversely larger $\rho$ permits more mass to be left unmatched. The optimal regularization $s_0$ is identified at the curve’s point of maximal positive curvature (the elbow), which  balances low transport cost against aggressive regularization.

In our discrete implementation, we sample $s$ uniformly from a sequence $\{s_i\}_{i=1}^{N}$ and compute the associated transportation costs $\zeta_i = \zeta(s_i)$. We compute the elbow by approximating the second derivative of the cost curve with respect to the regularization strength $\rho(s)$ by the centered second finite difference $\delta^2_\rho$,
\begin{equation*}
    \delta^2_{\rho}\zeta_i = \zeta_{i+1} - 2\zeta_i + \zeta_{i-1}\;\; \text{for}\;\; i = 2,\dots,N-1
\end{equation*}

and select the index with maximal positive curvature
\begin{equation*}
    i^{\star} = \argmax_{2\leq i\leq N-1}|\delta^2\zeta_i|,\quad  s_0 = s_{i^\star}
\end{equation*}
allowing us to analytically select the optimal regularization $s_0$. 

\subsection{Partial Soft-Matching as a Correlation Score}
Suppose that the tuning curves in two neuron populations $\bm{X}$ and $\bm{Y}$ have been mean-centered and scaled to unit-norm. Under this normalization, the inner product $\bm{x}_i^\top \bm{y}_j$ is identical to the Pearson correlation between neuron $i$ in $\bm{X}$ and neuron $j$ in $\bm{Y}$. Using this, the optimization can now be recast as a \emph{maximization} of total matched correlation:
\begin{equation*}
    d^{\mathrm{corr}}(\bm{X}, \bm{Y}) = \max_{\bm{T}\in\mathcal{T}^{s}(N_x, N_y)} \sum_{ij} \bm{T}_{ij}\bm{x}^\top_i \bm{y}_j
\end{equation*}
Intuitively, $d^{\mathrm{corr}}$ measures the average correlation between paired neurons under the coupling $\bm{T}$. We report $d^{\mathrm{corr}}$ for the remainder of the manuscript. We also report alignment obtained using a squared Euclidean cost function, $\bm{C}_{ij} = ||\bm{x}_i -\bm{x}_j||^2$, in Appendix~\ref{appendix: euclidean} and observe identical results.
%Intuitively, $d^{\mathrm{corr}}$ measures the average correlation between paired neurons under the optimal (partial) coupling $\bm{T}$. Because correlation-based scores are more immediately interpretable than abstract distance scalars due to their constrained range ($d^{\mathrm{corr}}\in [-1, 1]$), we report these correlation alignment scores for the remainder of the manuscript.

%Equivalently, one can derive conditions under which this is the same as minimizing a transport cost defined by a distance measure (Appendix~\ref{sec: eq-corr-dist}). 

\subsection{Interpretation and Output}
The optimal transport plan $\bm{T}^\star$ provides a soft partial alignment where:
\begin{itemize}
    \item Row sums $\in [0, 1/N_x]$: amount of mass transported from each source neuron
    \item Column sums $\in [0, 1/N_y]$: amount of mass received by each target neuron
    \item Total transported mass equals $s < 1$ (the fraction of total mass matched)
    \item Near-zero row/column sums identify effectively unmatched units
    %\item Maximum row/column sums $(1/N_x, 1/N_y)$ indicate units that contribute/receive their full capacity
\end{itemize}

This partitions populations based on participation in the optimal matching, from completely unmatched to maximally participating units.

% synthetic data
\section{Simulations: Robustness to Noise and Selecting the ``Correct'' Model}
\label{sec: synthetic}
We designed controlled simulations to evaluate whether partial soft-matching \textbf{(1)} maintains accurate correspondences despite spurious neurons and \textbf{(2)} correctly identifies which model shares more signal with a reference population. Synthetic neural representation generation is detailed in Appendix~\ref{sec: synth-generation}.
%Details for generating synthetic neural representation in Appendix~\ref{sec: synth-generation}.
%We designed controlled simulations to evaluate whether unbalanced soft-matching can \textbf{(1)} maintain accurate correspondences despite spurious neurons and \textbf{(2)} correctly identify which model shares more genuine signal with a reference population. We describe the procedure for generating synthetic neural representation in Appendix~\ref{sec: synth-generation}.
%Below we outline a toy experiment showing that unbalanced soft-matching is robust against spurious neurons---and can also help identify the ``correct'' model among two candidates. 

\subsection{Robustness Against Spurious Neurons}
We construct two neural populations $\bm{X}$ and $\bm{Y}$, each containing $K$ ``signal'' neurons matched pairwise. We introduce noise by augmenting $\bm{X}$ with $M_x$ random neurons and $\bm{Y}$ with $M_y$ random neurons, where each random neuron is drawn from $\varepsilon\sim\mathcal{N}(0, 1)$. The resultant populations are thus $\bm{X}\in\mathbb{R}^{(K+M_x)\times N}$ and $\bm{Y}\in\mathbb{R}^{(K+M_y)\times N}$, where $N$ is the number of unique stimuli.

\begin{figure}[htbp!]
    \centering
    \begin{tabular}{cc}
       \includegraphics[width=0.5\linewidth]{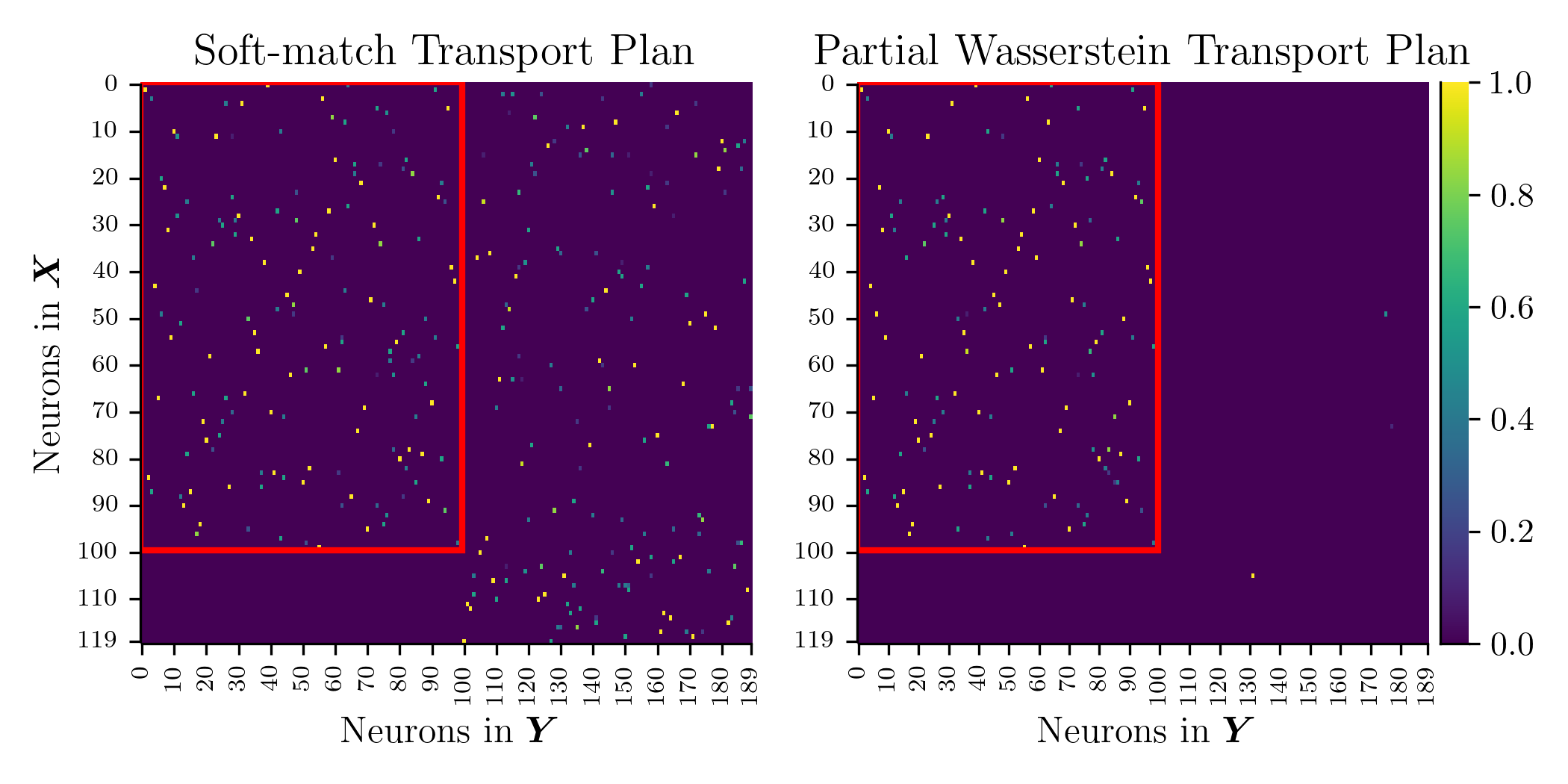} &  \includegraphics[width=0.3\linewidth]{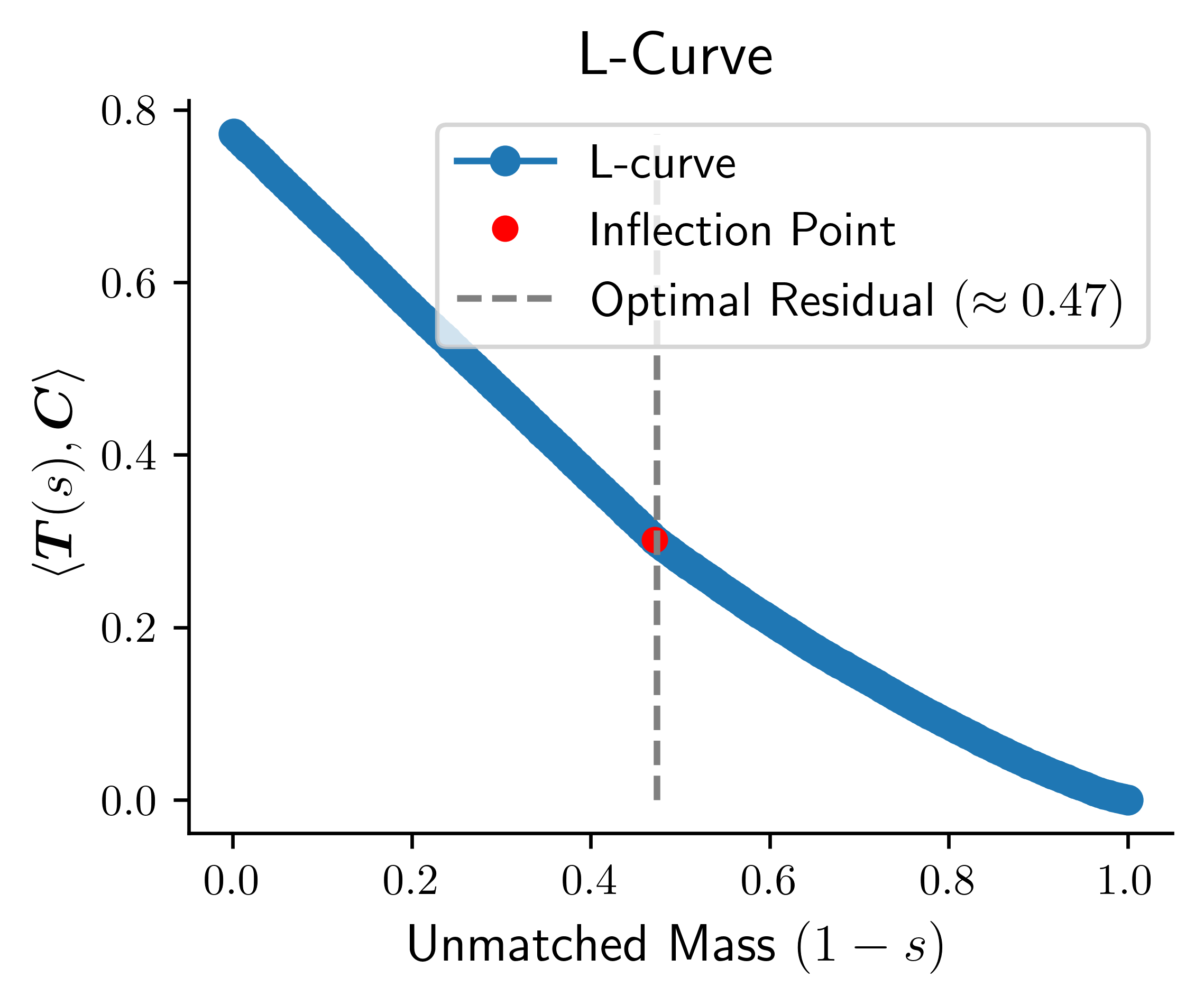}\\ 
    
       \textbf{(a)} & \textbf{(b)}
    \end{tabular}
    \caption{\textbf{Comparison of Balanced and Partial Soft-Matching.} \textbf{(a)} We simulate two neural representations, $\bm{X}$ and $\bm{Y}$, with 120 and 190 neurons respectively. The first 100 neurons represent pure \emph{signal}, while the rest are pure \emph{noise}. Red denotes all pure signal neurons in the two representations. \textbf{(b)} The L-curve method selects the optimal mass regularization parameter $(=90/190\approx 0.47)$, successfully discarding noisy units.
    }
    \label{fig:spurious-simulation}
\end{figure}

% \textbf{Comparison of Balanced and Unbalanced Soft-Matching.} \textbf{(a)} We simulate $2$ neural representations, $\bm{X}$ and $\bm{Y}$, having $120$ and $190$ neurons respectively. Of these, the first $100$ correspond to pure (matched) \emph{signal}, while the remaining represent pure \emph{noise}. \textbf{(b)} We use the L-curve method to precisely determine $(=90/190\approx 0.47)$ the optimal mass regularization parameter to discard noisy units.
In the (fully balanced) soft-matching distance must match \emph{all} $K+M_x$ neurons to $K+M_y$ neurons. This forces spurious outlier-to-outlier assignments, which inflate the overall transport cost and, consequently, the distance. In contrast, the partial soft-matching distance only transports mass corresponding to $K$ true matches, ignoring the random neurons. As a result, the recovered transport cost is significantly smaller and reflects the true correspondence between the signal neurons. We visualize the transport plans for both---soft-matching and partial soft-matching in Fig.~\ref{fig:spurious-simulation}, and observe that the L-curve heuristic is able to faithfully distinguish between noise and signal units.

\subsection{Choosing Between Two Models}
% Suppose we consider two models:
% \begin{enumerate}
%     \item \emph{Model A}: $\bm{Y}_a$ consists of a population of neurons that shares exactly $K$ correctly matched neurons with $\bm{X}$, along with an additional $M_y$ noisy neurons.
%     \item \emph{Model B}: $\bm{Y}_b$ is an alternative population that either \textbf{(i)} does not have the same signal neurons as $\bm{X}$, or \textbf{(ii)} shares fewer correctly matched neurons.
% \end{enumerate}
Suppose we consider two models: \emph{Model A}, where $\bm{Y}_a$ shares exactly $K$ correctly matched neurons with $\bm{X}$, along with $M_y$ additional noisy neurons; and \emph{Model B}, where $\bm{Y}_b$ \textbf{(i)} does not contain the same signal neurons as $\bm{X}$, and \textbf{(ii)} shares fewer correctly matched neurons.

\emph{Model A} is considered ``correct'' here because it preserves the maximum number of genuine signal correspondences with $\bm{X}$---the $K$ matched neurons encode the same computational features as their counterparts in $\bm{X}$---plus additional noisy neurons. These extra neurons may reflect measurement noise, inactive recording channels, or recording artifacts that are common in real neural data. \emph{Model B}, in contrast, shares only a subset of $\bm{X}'s$ signal neurons.

\begin{figure}[htbp!]
    \centering
    \begin{tabular}{cc}
      \includegraphics[width=0.35\linewidth]{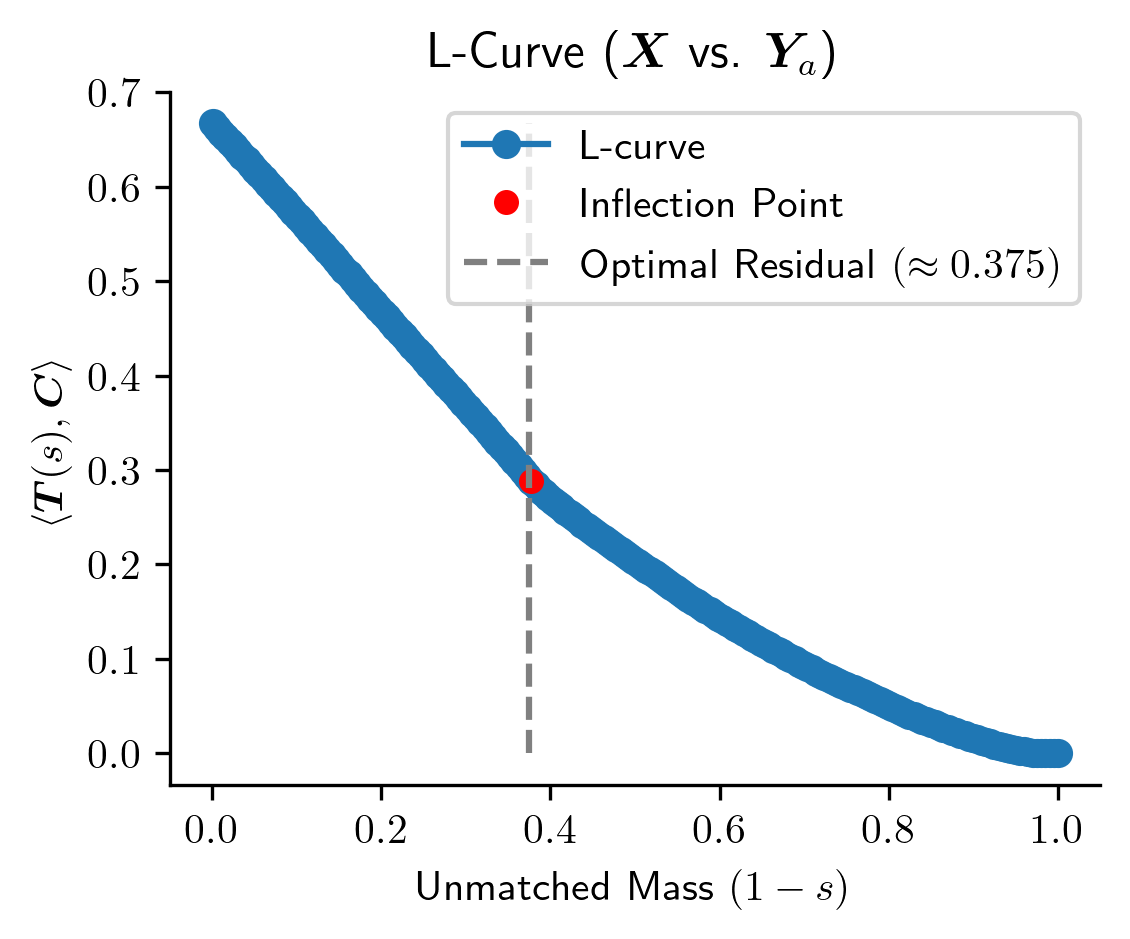} & \includegraphics[width=0.35\linewidth]{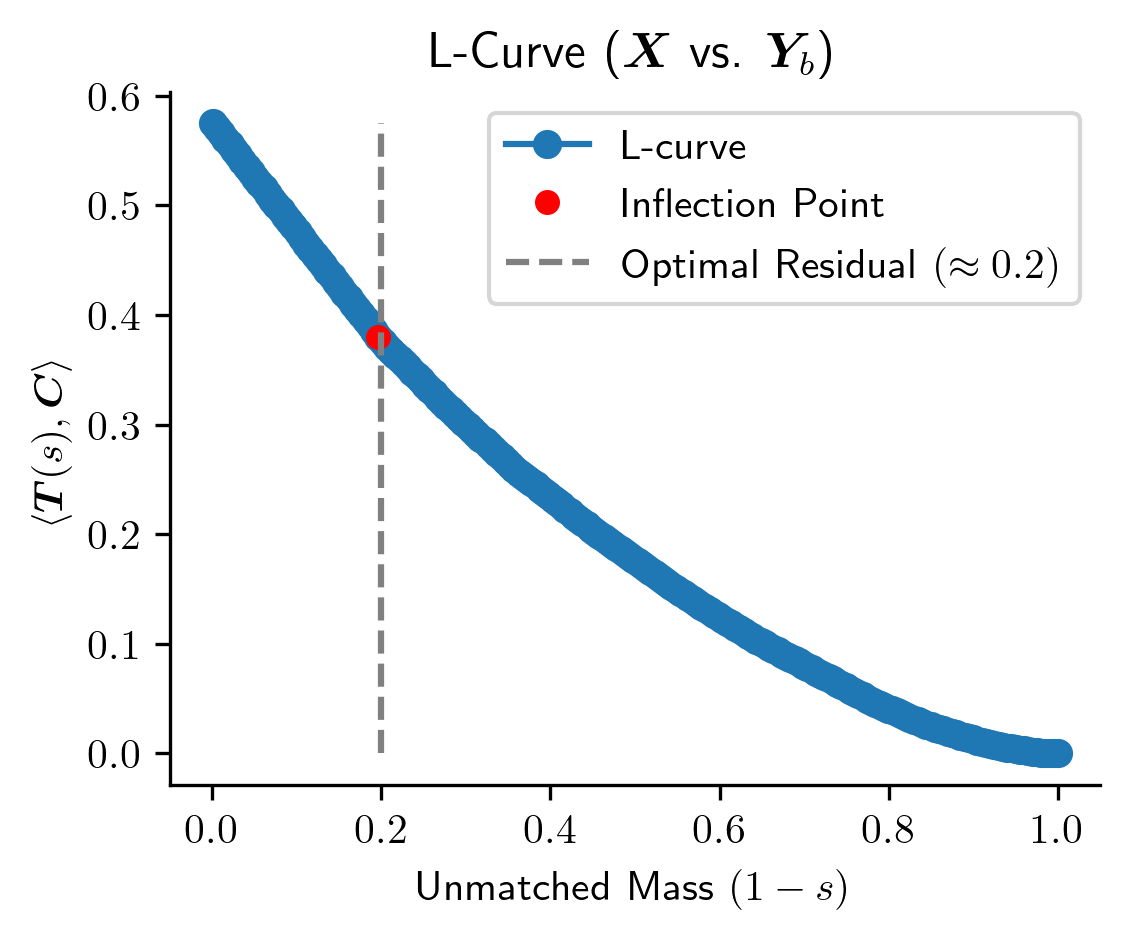}\\
    \textbf{(a)} & \textbf{(b)} 
    \end{tabular}
    \caption{ \textbf{Model Selection Using Partial Soft-Matching.} We simulate three synthetic representations to test whether partial soft-matching correctly identifies which of two candidate models---$\bm{Y}_a$ or $\bm{Y}_b$---shares more signal with a reference population $\bm{X}$ (100 units). $\bm{Y}_a$ contains all 100 signal units from $\bm{X}$ plus 60 noise units; $\bm{Y}_b$ contains 100 units, 80 of which match $\bm{X}$. The true fraction of shared units is known \emph{a priori}, marked by a vertical gray line. With the L-curve-selected regularization, partial soft-matching yields correlation scores $s_{\mathrm{partial}}(\bm{X}, \bm{Y}_a)=0.715$ and $s_{\mathrm{partial}}(\bm{X}, \bm{Y}_b)=0.645$, correctly favoring $\bm{Y}_a$. Standard soft-matching fails, with $s_{\mathrm{sm}}(\bm{X}, \bm{Y}_a) = 0.339$ and $s_{\mathrm{sm}}(\bm{X}, \bm{Y}_b) = 0.415$, incorrectly preferring $\bm{Y}_b$ due to forced matching of noise.}

    % \textbf{Model Selection Using Unbalanced Soft-Matching.} We simulate three synthetic neural representations and evaluate whether unbalanced soft-matching correctly identifies which of two candidate models---$\bm{Y}_a$ and $\bm{Y}_b$---shares greater signal with a reference population $\bm{X}$ ($100$ units).
    % Representation $\bm{Y}_a$ contains all $100$ signal units from $\bm{X}$ plus $60$ additional noise units, while $\bm{Y}_b$ contains $100$ units, $80$ of which share signal with $\bm{X}$ ($20$ unmatched). Since the true number of shared units is known \emph{a priori}, the corresponding optimal residual is marked by a vertical gray line. Using the regularization parameter selected via the L-curve, we compute unbalanced correlation scores $s_{\mathrm{partial}}(\bm{X}, \bm{Y}_a)=0.715$ and $s_{\mathrm{partial}}(\bm{X}, \bm{Y}_b)=0.645$, which correctly identifies $\bm{Y}_a$ as the model with higher proportion of shared signal. On the other hand, soft-matching fails to distinguish the models correctly, yielding $s_{\mathrm{sm}}(\bm{X}, \bm{Y}_a) = 0.339$ and $s_{\mathrm{sm}}(\bm{X}, \bm{Y}_b) = 0.415$, incorrectly favoring model $\bm{Y}_b$, demonstrating how forced matching of noise corrupts the similarity assessment.
    %}
    \label{fig: model-selection}
\end{figure}    

We compute the partial soft-matching scores $s_{\textrm{partial}}(\bm{X}, \bm{Y}_a)$ and $s_{\textrm{partial}}(\bm{X}, \bm{Y}_b)$. Because partial OT can ignore outliers and preserve only the true $K$ matches, the distance, to the \emph{``correct''} model $\bm{Y}_a$ will be significantly smaller---equivalently, the correlation satisfies $s_{\textrm{partial}}(\bm{X}, \bm{Y}_a) > s_{\textrm{partial}}(\bm{X}, \bm{Y}_b)$, correctly identifying \emph{Model A} as sharing more signal with $\bm{X}$. By contrast, standard soft-matching forces matches for all units (including noise), obscuring signal differences and failing to discriminate $\bm{Y}_a$ from $\bm{Y}_b$, as shown in Fig.~\ref{fig: model-selection}.
%We compute the unbalancled soft-matching scores $s_{\textrm{partial}}(\bm{X}, \bm{Y}_a)$ and $s_{\textrm{partial}}(\bm{X}, \bm{Y}_b)$. Because partial OT can ignore outliers and preserve only the true $K$ matches, the distance, to the \emph{``correct''} model $\bm{Y}_a$ will be significantly smaller. Alternatively, in the correlation-based formulation we observe a larger correlation with the correct model $\bm{Y}_a$, i.e.: $s_{\textrm{partial}}(\bm{X}, \bm{Y}_a) > s_{\textrm{partial}}(\bm{X}, \bm{Y}_b)$.  This correctly identifies \emph{Model A} as having greater shared signal with $\bm{X}$. In contrast, standard soft-matching, which must match \emph{all} units including noise, fails to reliably discriminate between $\bm{Y}_a$ and $\bm{Y}_b$, as shown in Fig.~\ref{fig: model-selection}. The forced matching of numerous noisy units obscures the underlying difference in signal correspondence, yielding misleading similarity scores.

%In contrast, the \emph{fully balanced} or \emph{soft-matching distance} fails to discriminate meaningfully between $\bm{Y}_a$ and $\bm{Y}_b$ when both contain a substantial number of unmatched or noisy units. 

% NSD + geirhos
\section{Applications in Neuroscience and AI}
\label{sec:applications}

\subsection{Comparisons of Neural Recordings Across Subjects}
\begin{figure}[htbp!]
    \centering
    \includegraphics[width=0.8\linewidth]{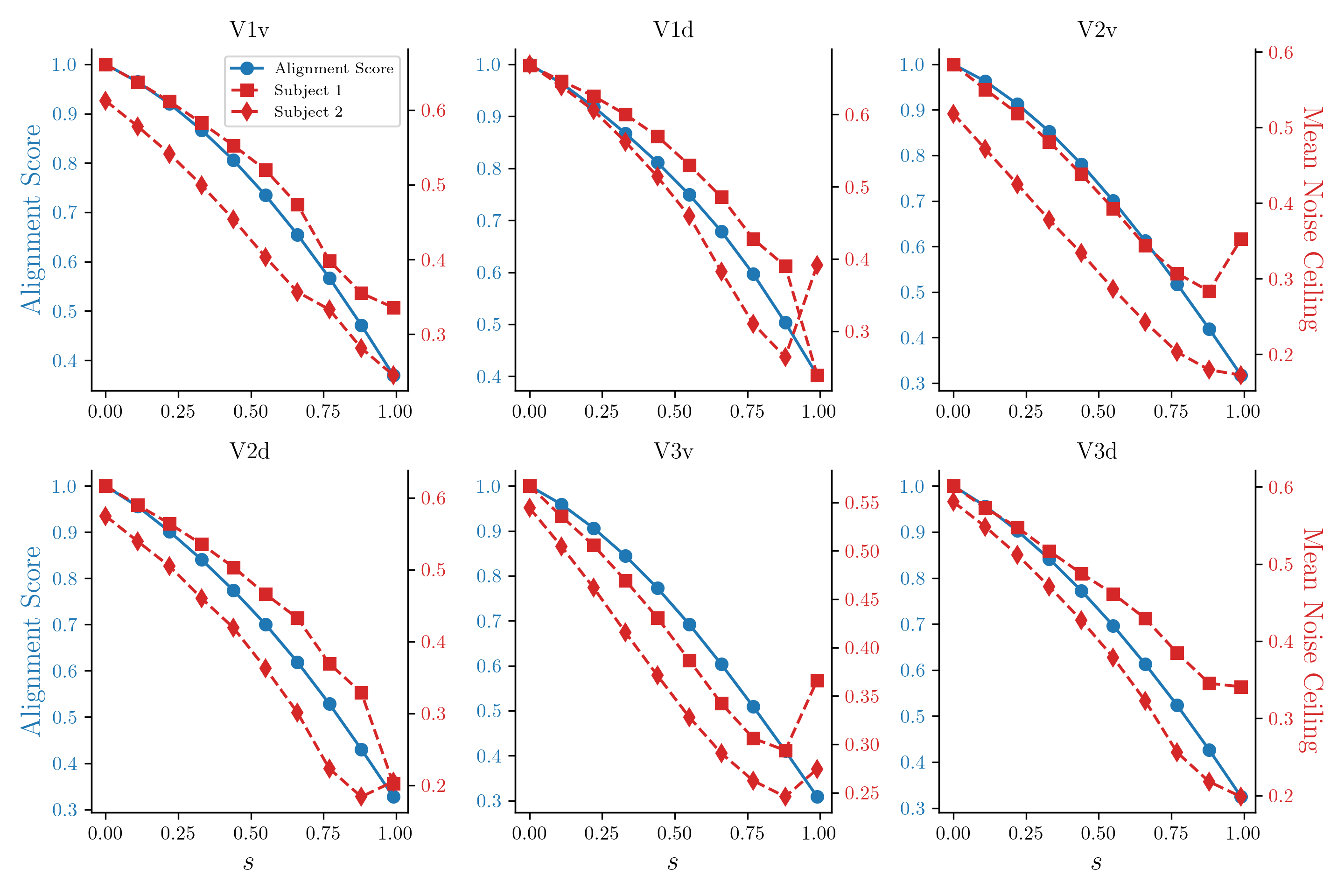}
    \caption{\textbf{Aligning Voxel Responses Between Different Subjects in NSD.} 
    For each area, we plot the \textbf{(i)} partial soft-matching score at different mass regularization values and \textbf{(ii)} the mean noise ceilings of the voxels that were kept at that regularization. The alignment criterion consistently identifies low noise-ceiling voxels for exclusion.}
    \label{fig:nsd-noise-ceiling}
\end{figure}
Perfect correspondence of neural populations across subjects is rare---measurement noise, inactive voxels and anatomical variability implies imprecise region boundary definition. Voxels nominally assigned to the same brain area can sample neighboring regions implementing distinct computations. Individual differences in functional organization can further aggravate distinct computations across subjects. Partial soft-matching addresses these challenges by selectively excluding non-corresponding units from the alignment.
%When comparing neural populations across subjects, perfect correspondence is rarely achievable. Beyond measurement noise and inactive voxels, anatomical variability means that region boundaries are imprecisely defined---voxels nominally assigned to the same brain area may actually sample from neighboring regions implementing distinct computations. Additionally, individual differences in functional organization mean that some computations may be expressed in one subject but not another. Unbalanced soft-matching addresses these challenges by selectively excluding non-corresponding units from the alignment.

We demonstrate this on voxel responses from a subject pair (IDs $1$ and $2$) across six visual areas $\mathrm{(V1v, V1d, V2v, V2d, V3v, V3d)}$ from the Natural Scenes Dataset~\citep{allen2022massive}. Fig.~\ref{fig:nsd-noise-ceiling} shows how voxel selection quality changes as we vary the mass regularization parameter $s$ from $1$ (including all voxels) to $0$ (excluding all voxels). As $s$ decreases and we exclude more voxels, the mean noise ceiling of the retained voxels steadily increases, while the alignment score between these retained voxels also improves. Since noise ceiling measures the reliability of a voxel's responses across repeated stimulus presentations, this demonstrates that our method successfully identifies and excludes voxels with poor response replicability. By progressively discarding these unreliable measurements, partial soft-matching automatically focuses the alignment on the subset of voxels that provide the most consistent and well-matched signal across subjects. We perform an identical experiment on a different NSD subject pair (Appendix~\ref{sec: noise-ceiling-nsd-alt}) and observe identical results.

\begin{wrapfigure}{R!}{0.6\columnwidth}
    \begin{center}
        \includegraphics[width=0.58\columnwidth]{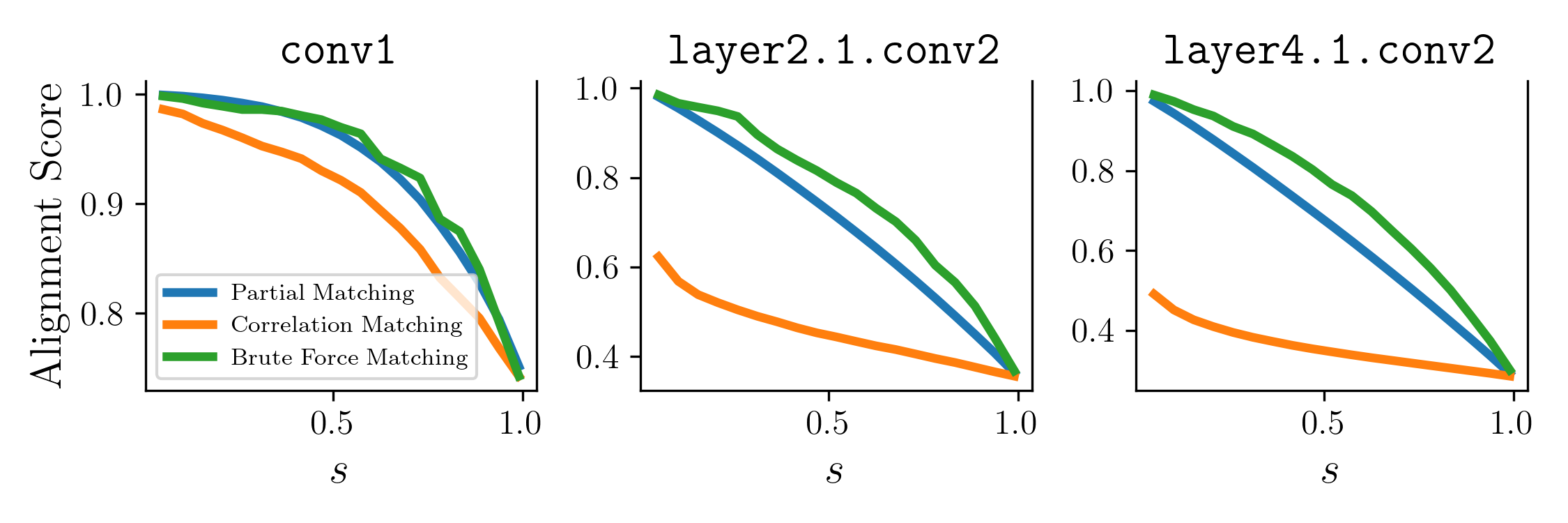}
        \caption{\textbf{Evaluating Methods for Identifying (Un)matched Neurons in Deep Networks.} We compare three methods for ranking convolutional kernels by alignment between two ResNet-$18$ models trained from different random initializations on ImageNet, across early, middle, and late layers. Removing low-alignment units identified by partial soft-matching yields alignment scores nearly identical to those obtained by removing kernels ranked least important via brute-force ablations, while correlation-based rankings perform poorly.}%We evaluate three methods for ranking convolutional kernels by their degree of alignment between two ResNet-$18$ models trained from different random initializations on ImageNet, testing at early, middle, and late convolutional layers. When using these rankings to identify and remove poorly-aligned kernels, our unbalanced soft-matching method achieves alignment scores nearly identical to computationally expensive brute-force matching across all layer depths, while correlation-based ordering fails to correctly identify which kernels matter for alignment.}
        \label{fig:resnet18-all-baselines}
    \end{center}
\end{wrapfigure}
\subsection{Comparison Against Baseline Methods}
\label{sec: baseline-comparison}
\begin{figure}[htbp!]
    \centering
    \includegraphics[width=0.8\linewidth]{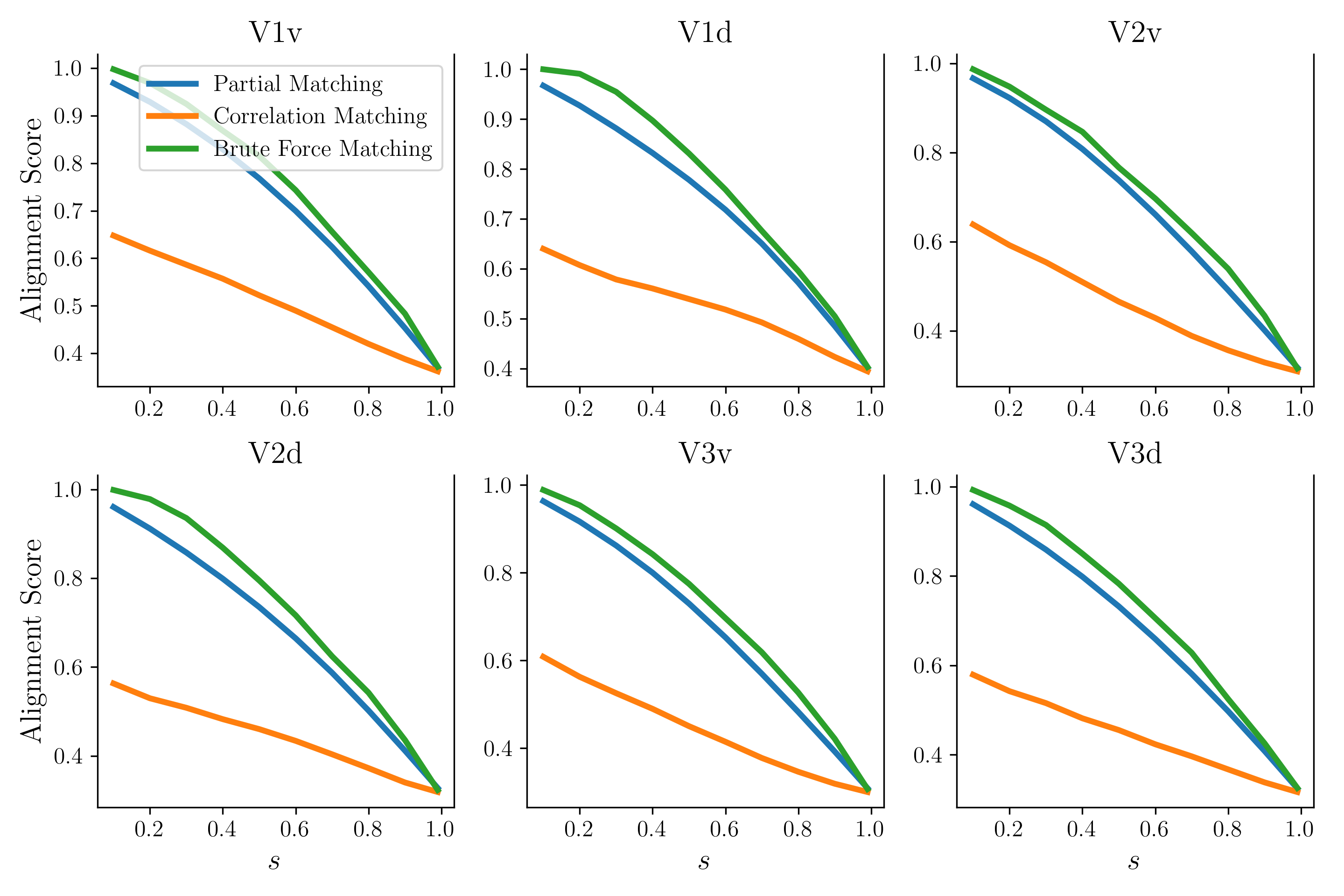}
    \caption{\textbf{Evaluating Methods for Identifying (Un)matched Voxels in Brain Data.} We evaluate three methods for ranking voxels by their degree of alignment between a subject pair from NSD across six visual areas. Removing low-alignment voxels identified by partial soft-matching yields alignment scores nearly identical to those obtained by removing voxels ranked least important via brute-force ablations, while correlation-based rankings perform poorly.  
    }
    \label{fig:nsd-all-baselines}
\end{figure}

In this section, we demonstrate the utility of our metric as an efficient tool for rank-ordering neurons by their degree of cross-population alignment. We compare three approaches---brute-force matching, correlation-based ordering and our proposed partial soft-matching method. We test these methods in two distinct settings: \textbf{(1)} comparing convolutional kernels between two ResNet-$18$ models trained from different random initializations on ImageNet~\citep{5206848}, examining early, middle, and late layers (Fig.~\ref{fig:resnet18-all-baselines}), and \textbf{(2)} aligning voxel responses between human subjects viewing natural images, across six visual areas $\mathrm{(V1v, V1d, V2v, V2d, V3v, V3d)}$ from the Natural Scenes Dataset (Fig.~\ref{fig:nsd-all-baselines}).

\emph{Brute-force matching} provides the ground-truth ranking by exhaustively testing each neuron's contribution to alignment. We fit an optimal soft-matching transformation to the complete representation, then iteratively remove each neuron and recompute the entire soft-matching optimization to measure the impact on alignment score. This produces an exact ranking of neurons by their alignment quality. However, each soft-matching optimization requires $\mathcal{O}(n^3\log n)$ operations and for $n$ neurons to test, the total complexity is $\mathcal{O}(n^4\log n)$, making this approach computationally prohibitive for realistic population sizes.

\emph{Correlation-based ordering} attempts a computationally cheaper approximation by computing pairwise Pearson correlations between neurons using the transport plan from a single soft-matching optimization. As shown in Figures~\ref{fig:resnet18-all-baselines} and ~\ref{fig:nsd-all-baselines}, this heuristic fails catastrophically---it incorrectly identifies and removes neurons that are actually crucial for alignment, resulting in dramatically degraded alignment scores. This failure occurs because individual correlation values don't capture the global optimization structure of the transport problem.

\emph{Partial soft-matching} offers a nuanced tradeoff. To obtain a complete ranking of all $n$ neurons (matching the output of brute-force), we would still require $n$ separate optimizations at different regularization values, maintaining $\mathcal{O}(n^4\log n)$ complexity. However, for the practically relevant task of identifying the top $X\%$ most-aligned or least-aligned neurons---which suffices for most analyses in neuroscience and deep learning which require identifying highly-aligned or poorly-aligned subpopulations rather than complete rankings---a single optimization at the appropriate regularization value $\left(\mathcal{O}(n^3\log n)\right)$ provides near-identical results to brute-force ranking. As Figures~\ref{fig:resnet18-all-baselines} and ~\ref{fig:nsd-all-baselines} demonstrate, when selecting subsets of neurons at various alignment thresholds, our method's selections yield alignment scores nearly matching those from exhaustive brute-force ranking, while correlation-based selection performs poorly. Full algorithmic details are provided in Appendix~\ref{sec: baseline-matching-algos}. 

\subsection{Mapping (Dis)similar Brain Regions}
A robust similarity metric for neural populations should exhibit specificity: it must identify when responses come from the same brain area across subjects (true positives) while avoiding false matches between distinct areas (false positives)~\citep{thobani2025model}. This specificity is crucial when anatomical boundaries are imprecise and individual variability is high.

We evaluate this specificity by testing whether partial soft-matching correctly aligns homologous visual areas while maintaining separation between distinct areas. Concretely, we select two visual ROIs in a subject pair and then compute a between-subject matching for all voxels of these regions. For each pair of visual regions within and across subjects from the NSD, we compute the precision of voxel assignments---the fraction of matched voxels that truly belong to corresponding regions. Table~\ref{tab:region-mapping} shows these precision scores comparing standard soft-matching (which must match all voxels), thresholding using voxel noise ceilings, and partial soft-matching (which can exclude poor correspondences). The optimal regularization parameter for partial soft-matching is chosen via the L-curve heuristic as described in Section~\ref{sec: optimal-mreg}.

Across most region pairs, partial soft-matching achieves higher precision than standard soft-matching and thresholding, with particularly striking improvements for several cross-area comparisons (\emph{e.g.,} $\mathrm{V1d} + \mathrm{V2v}$: $0.906 \rightarrow 0.971$). This improvement stems from the method's ability to exclude voxels that lack clear correspondence—whether due to boundary uncertainty or measurement noise. By not forcing these ambiguous voxels into the matching, partial soft-matching maintains cleaner separation between distinct regions while preserving strong alignment within homologous areas.  

%\begin{SCtable}
\begin{table}[htbp!]
%\begin{wraptable}{r}{0.6\columnwidth}
    \centering
    %\resizebox{ \linewidth}{!}{
    \begin{tabular}{ccccc}
    \toprule
    Brain Region Pair & SM Precision $(\uparrow)$ & ParSM Precision $(\uparrow)$ & $\epsilon=0.1\; (\uparrow)$ & $\epsilon=0.3\; (\uparrow)$\\
    \midrule
    $\mathrm{V1v + V1d}$ & $0.839$ & $\mathbf{0.905}\; (0.76)$ & $0.847\; (0.98)$ & $0.855\; (0.94)$\\
    $\mathrm{V1v + V2v}$ & $0.677$ & $0.680\; (0.99)$          & $0.680\; (0.96)$ & $\mathbf{0.695}\; (0.88)$\\
    $\mathrm{V1v + V2d}$ & $0.880$ & $0.884\; (0.99)$          & $0.884\; (0.97)$ & $\mathbf{0.894}\; (0.91)$\\
    $\mathrm{V1v + V3v}$ & $0.798$ & $\mathbf{0.853}\; (0.97)$ & $0.803\; (0.99)$ & $0.815\; (0.90)$\\
    $\mathrm{V1v + V3d}$ & $0.882$ & $0.890\; (0.98)$          & $0.890\; (0.97)$ & $\mathbf{0.913}\; (0.91)$\\
    $\mathrm{V1d + V2v}$ & $0.881$ & $\mathbf{0.971}\; (0.71)$ & $0.889\; (0.96)$ & $0.906\; (0.89)$\\
    $\mathrm{V1d + V2d}$ & $0.706$ & $0.708\; (0.99)$          & $0.720\; (0.97)$ & $\mathbf{0.727}\; (0.92)$\\
    $\mathrm{V1d + V3v}$ & $0.879$ & $0.881\; (0.99)$          & $0.885\; (0.97)$ & $\mathbf{0.892}\; (0.91)$\\
    $\mathrm{V1d + V3d}$ & $0.803$ & $\mathbf{0.878}\; (0.76)$ & $0.818\; (0.97)$ & $0.828\; (0.92)$\\
    $\mathrm{V2v + V2d}$ & $0.869$ & $0.879\; (0.98)$          & $0.880\; (0.95)$ & $\mathbf{0.896}\; (0.87)$\\
    $\mathrm{V2v + V3v}$ & $0.651$ & $\mathbf{0.661}\; (0.95)$ & $0.653\; (0.95)$ & $0.654\; (0.86)$\\
    $\mathrm{V2v + V3d}$ & $0.853$ & $0.856\; (0.99)$          & $0.867\; (0.95)$ & $\mathbf{0.882}\; (0.87)$\\
    $\mathrm{V2d + V3v}$ & $0.833$ & $\mathbf{0.971}\; (0.42)$ & $0.845\; (0.96)$ & $0.863\; (0.88)$\\
    $\mathrm{V2d + V3d}$ & $0.638$ & $\mathbf{0.643}\; (0.99)$ & $0.642\; (0.96)$ & $0.643\; (0.89)$\\
    $\mathrm{V3v + V3d}$ & $0.814$ & $0.822\; (0.99)$          & $0.828\; (0.96)$ & $\mathbf{0.852}\; (0.88)$\\
    \bottomrule
    \end{tabular}
    %}
    %\begin{tabular}{ccc}
    %\toprule
    %     Brain Region Pair & SM Precision $(\uparrow)$ & UnSM Precision $(\uparrow)$ \\
    %     \midrule
    %     $\mathrm{V1v + V1d}$ & $0.839$ & $\mathbf{0.905}$\\
    %     $\mathrm{V1v + V2v}$ & $0.677$ & $\mathbf{0.680}$\\
    %     $\mathrm{V1v + V2d}$ & $0.880$ & $\mathbf{0.884}$\\
    %     $\mathrm{V1v + V3v}$ & $0.798$ & $\mathbf{0.853}$\\
    %     $\mathrm{V1v + V3d}$ & $0.882$ & $\mathbf{0.890}$\\
    %     $\mathrm{V1d + V2v}$ & $0.881$ & $\mathbf{0.971}$\\
    %     $\mathrm{V1d + V2d}$ & $0.706$ & $\mathbf{0.708}$\\
    %     $\mathrm{V1d + V3v}$ & $0.879$ & $\mathbf{0.881}$\\
    %     $\mathrm{V1d + V3d}$ & $0.803$ & $\mathbf{0.878}$\\
    %     $\mathrm{V2v + V2d}$ & $0.869$ & $\mathbf{0.879}$\\
    %     $\mathrm{V2v + V3v}$ & $0.651$ & $\mathbf{0.661}$\\
    %     $\mathrm{V2v + V3d}$ & $0.853$ & $\mathbf{0.856}$\\
    %     $\mathrm{V2d + V3v}$ & $0.833$ & $\mathbf{0.971}$\\
    %     $\mathrm{V2d + V3d}$ & $0.638$ & $\mathbf{0.643}$\\
    %     $\mathrm{V3v + V3d}$ & $0.814$ & $\mathbf{0.822}$\\
    %     \bottomrule
    %\end{tabular}
    %}
    \caption{\textbf{Precision of Cross-Subject Voxel Alignment Within and Across Visual Areas}. Comparison of soft-matching (SM), partial soft-matching (ParSM) and noise ceiling thresholding to align voxels between visual regions in an NSD subject pair. Precision measures the fraction of matched voxels belonging to corresponding anatomical regions (higher = better specificity). We include the fraction of total voxels that contribute towards computing alignment in parenthesis. The $\epsilon$ values denote the noise ceiling threshold below which voxels are excluded. ParSM almost always yields higher precision by excluding voxels that lack clear correspondence.}
    \label{tab:region-mapping}
%\end{wraptable}
\end{table}
%\end{SCtable}

%Unbalanced soft-matching consistently achieves superior precision by excluding voxels lacking clear correspondence.

\subsection{Maximally Exciting Images}
Maximally Exciting Images (MEIs)---synthetic stimuli optimized to maximize individual unit responses---provide an interpretable visualization of what each neuron ``looks for'' in its input~\citep{erhan2009visualizing, pierzchlewicz2023energy, walker2019inception, bashivan2019neural}. We synthesize MEIs\footnote{Appendix~\ref{sec: mei-synthesis}} for unit pairs from two ResNet-18 models trained with different random seeds, sampling from neurons ranked as highly-matched (top $10\%$ of transport mass) versus poorly-matched (bottom $10\%$) by our metric. Fig.~\ref{fig: mei-matching} shows striking differences: highly-matched pairs produce nearly identical MEIs, revealing that these units have converged on similar feature detectors despite independent training. In contrast, unmatched pairs yield divergent MEIs with distinct visual patterns, confirming they likely implement different computations. We demonstrate this with some additional results in Appendix~\ref{sec: mei-extra}.

%We use MEIs to validate whether our metric correctly identifies matched vs. unmatched units across networks \textcolor{red}{(Fig.~\ref{fig: mei-matching})}.
%\begin{figure}[htbp!]
%    \centering
%    \includegraphics[width=\columnwidth]{./figures/dnn/mei-table.png}
%    \caption{\textbf{Visualization of (Un)matched Units Using Maximally Exciting Images.} We show MEIs for three layers of ResNet$18$, comparing units classified as matched or unmatched by the unmatched soft-matching metric. Each MEI was optimized to maximize the response at the center spatial location of a given feature-map channel in a pair of ResNet$18$ models that differ only in random seed. Matched examples are sampled from the top $10\%$ of unbalanced soft-matching scores; unmatched examples are sampled from the bottom $10\%$. Note that as depth increases, finding nearly identical MEIs between the two models becomes increasingly difficult.}
%    \label{fig: mei-matching}
%end{figure}
\newlength{\imgheight}
\setlength{\imgheight}{1.75cm}
\begin{center}
\begin{tabular}{@{} p{0.47\textwidth} | p{0.47\textwidth} @{}}

% column headers
\multicolumn{1}{@{}p{0.47\textwidth}@{}}{\centering{\textbf{\fontfamily{cmr}\selectfont Matched}}} &
\multicolumn{1}{@{}p{0.47\textwidth}@{}}{\centering{\textbf{\fontfamily{cmr}\selectfont Unmatched}}} \\[4pt]

% row 1 
\centering
% left column
\includegraphics[height=\imgheight, keepaspectratio]{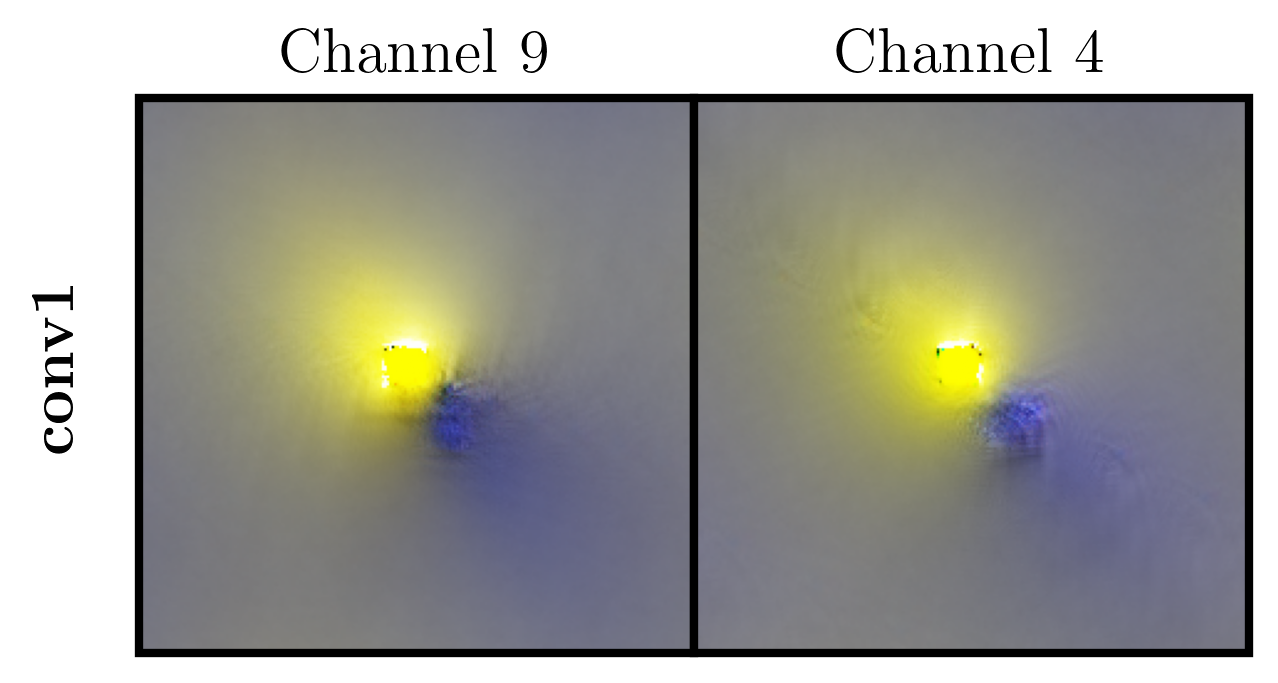}\hspace{0.01\linewidth} 
\includegraphics[height=\imgheight, keepaspectratio]{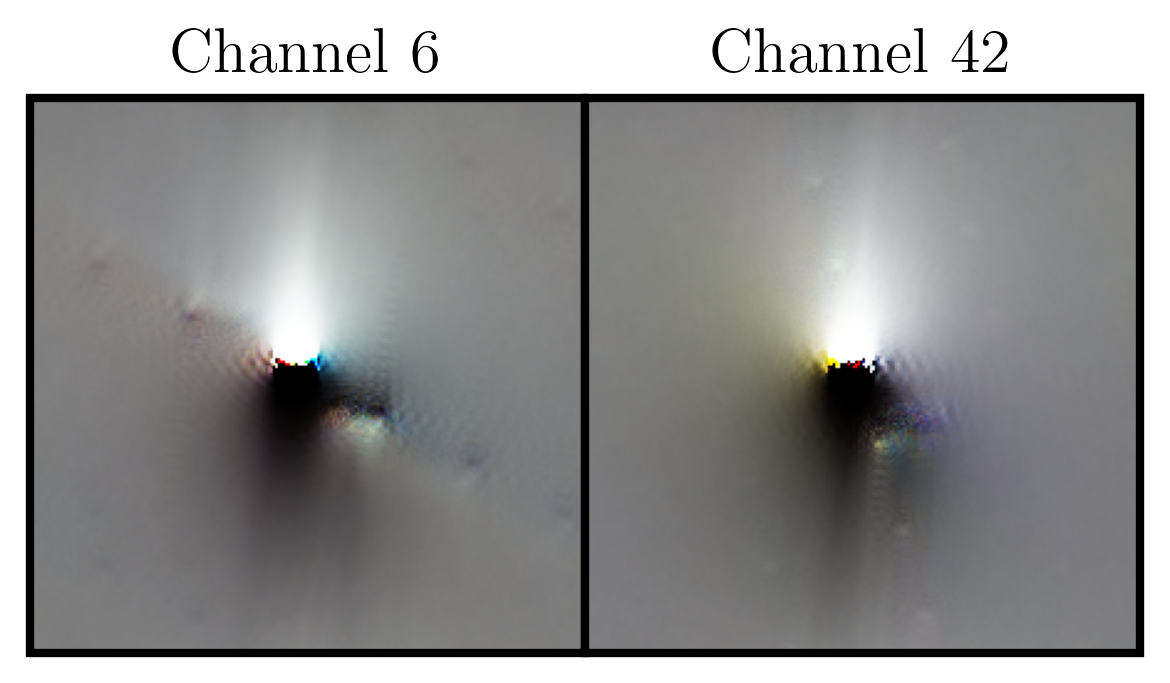}
&
% right outer column
\includegraphics[height=\imgheight, keepaspectratio]{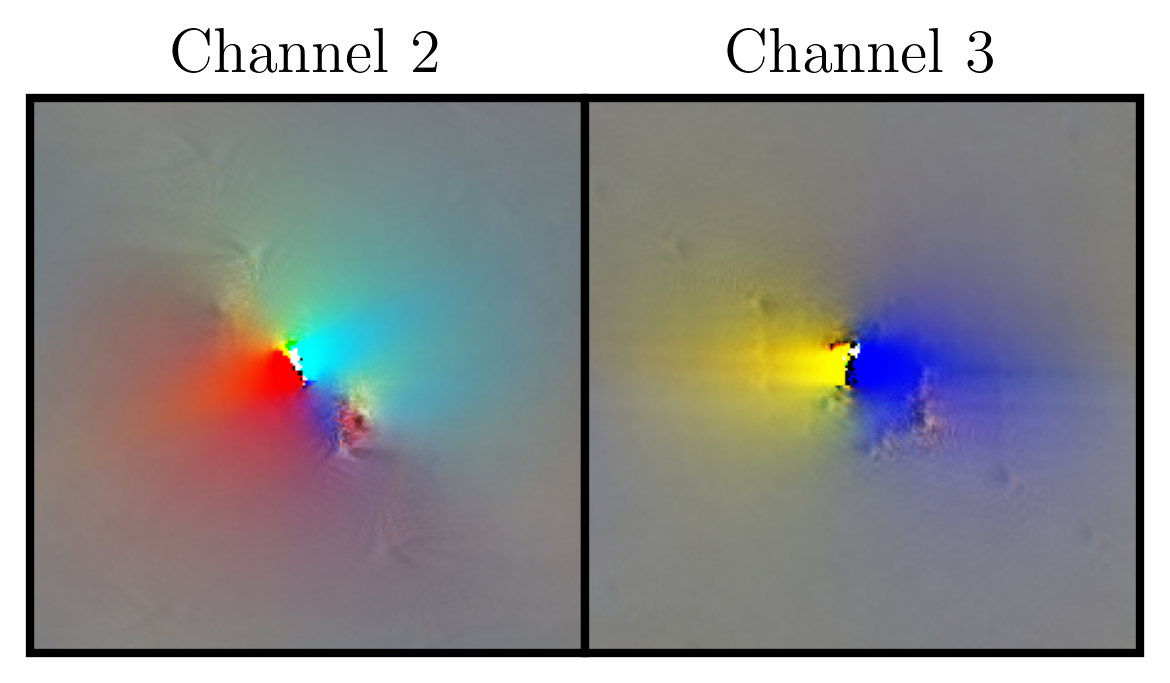}\hspace{0.01\linewidth}%
\includegraphics[height=\imgheight, keepaspectratio]{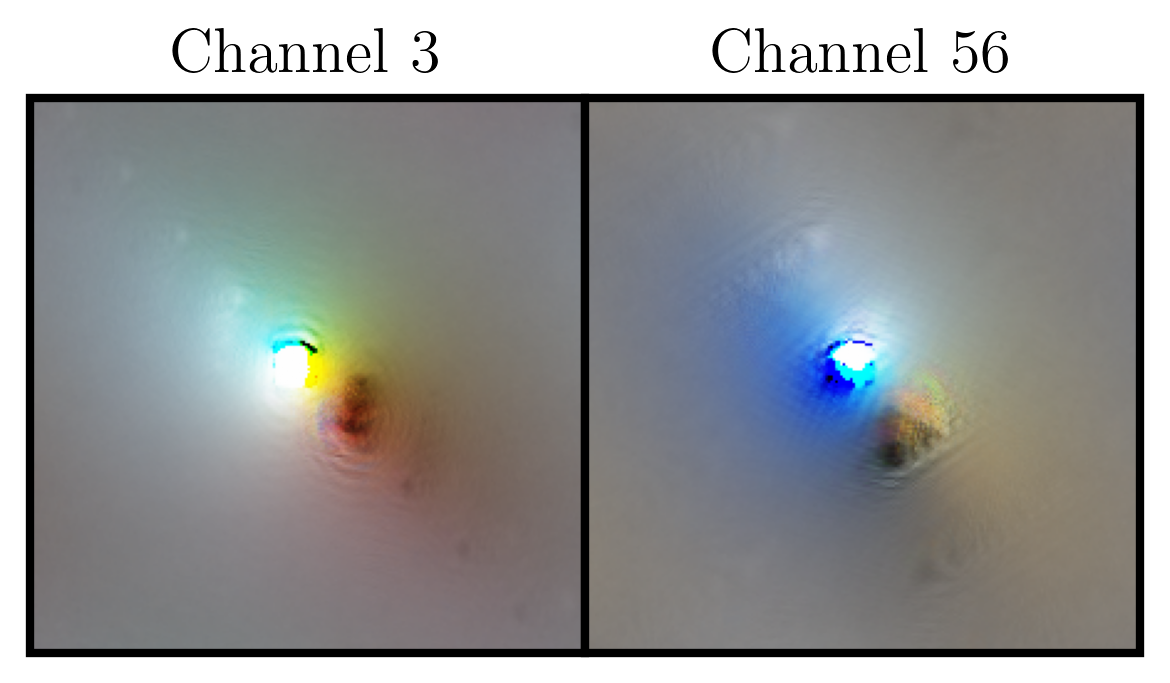}
\\[1pt]

% row 2 
\centering
\includegraphics[height=\imgheight, keepaspectratio]{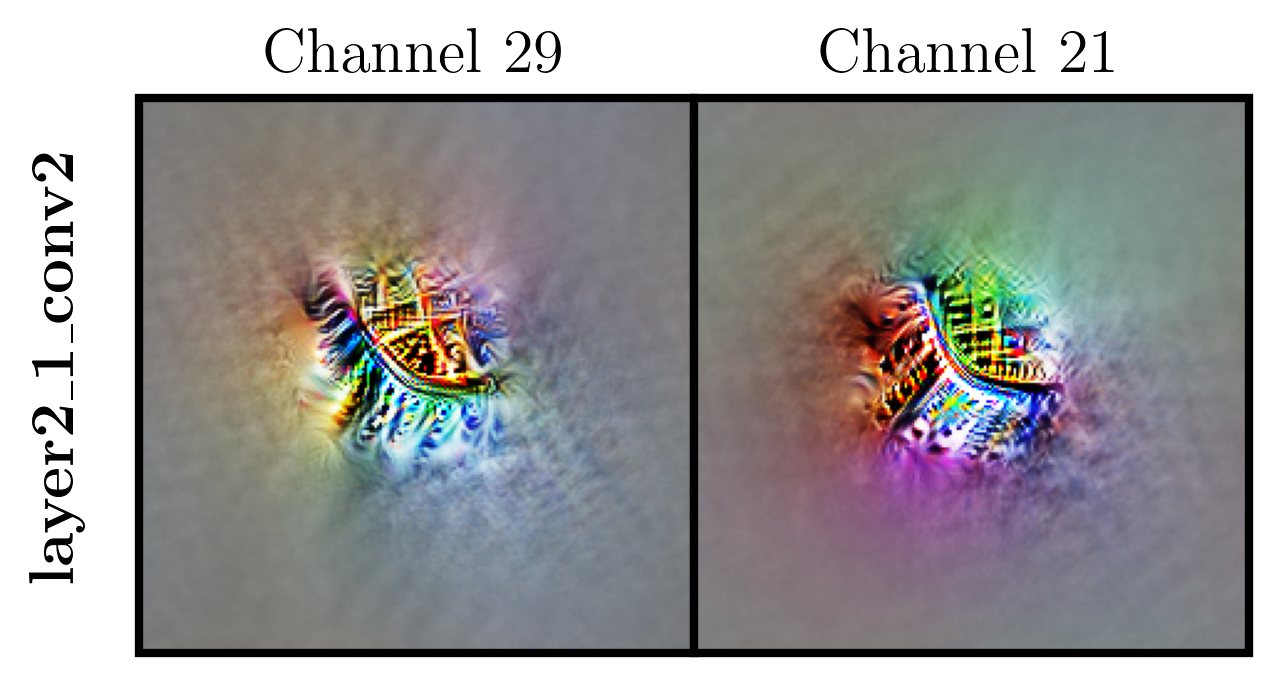}\hspace{0.01\linewidth}%
\includegraphics[height=\imgheight, keepaspectratio]{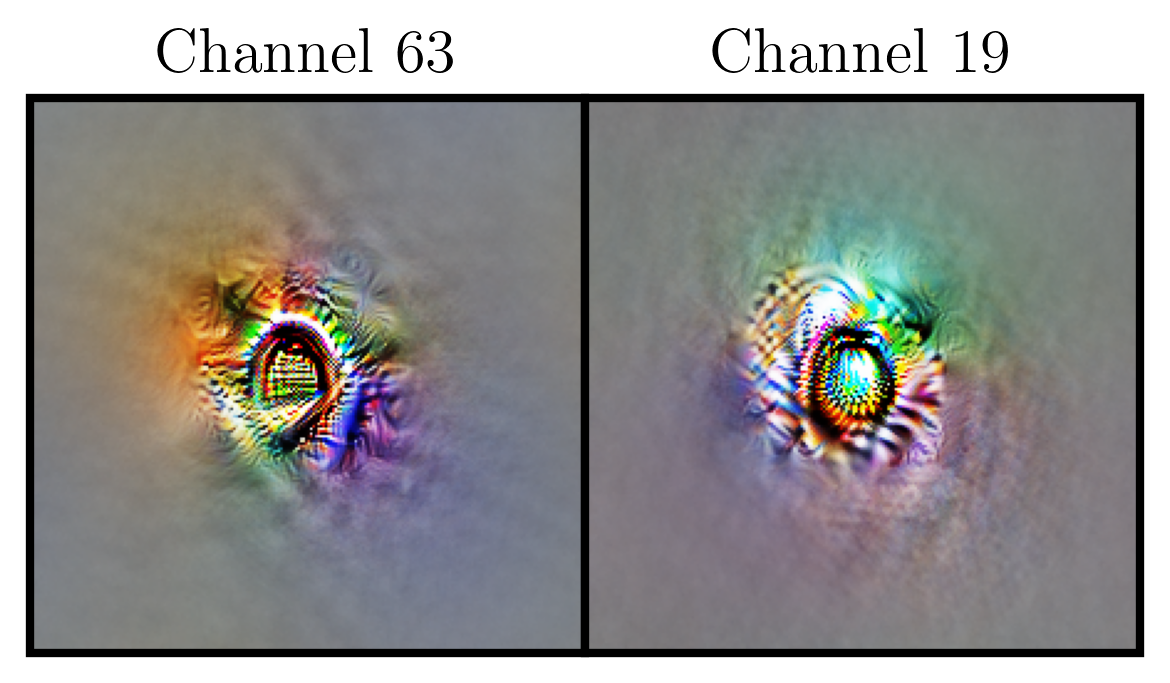}
&
\includegraphics[height=\imgheight, keepaspectratio]{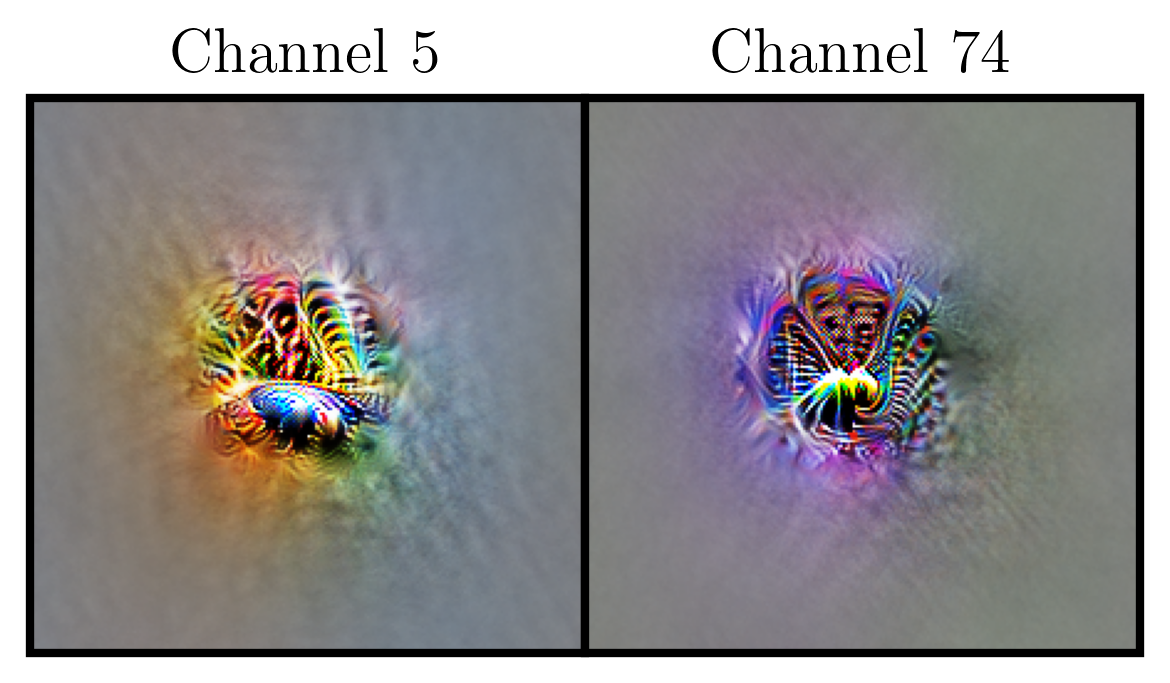}\hspace{0.01\linewidth}%
\includegraphics[height=\imgheight, keepaspectratio]{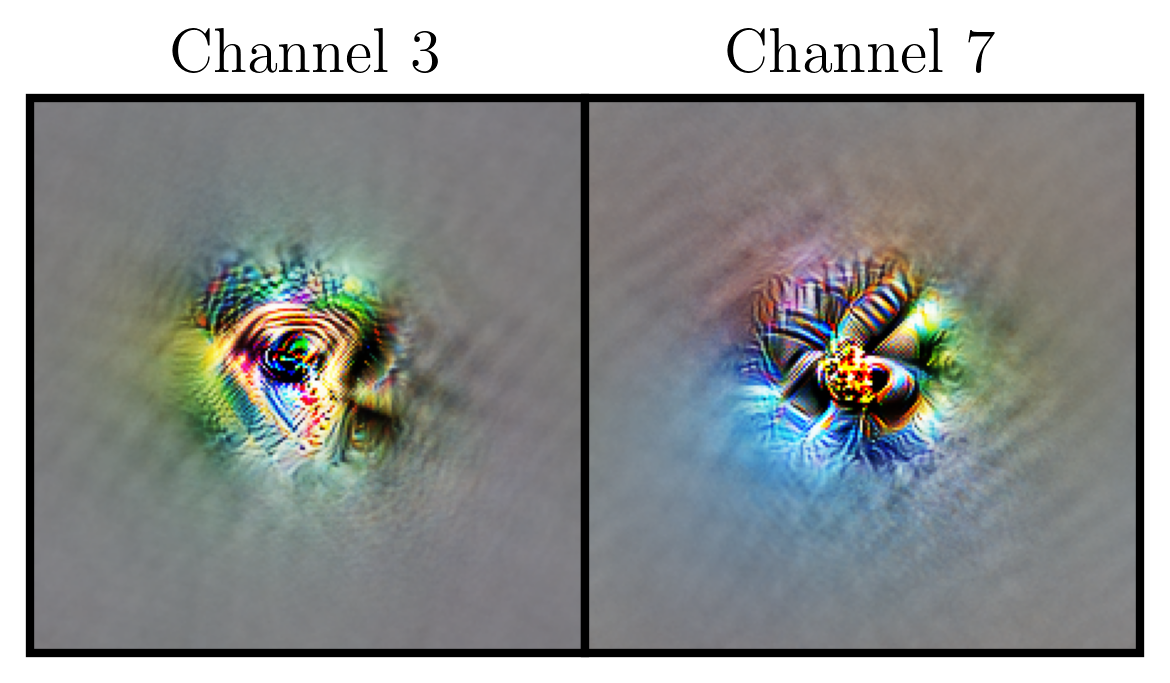}
\\%[1pt]

%% row 3 
%\centering
%\includegraphics[height=\imgheight, keepaspectratio]{./figures/dnn/mei-matches/matched/layer4_1_conv2_210_91.png}\hspace{0.01\linewidth}%
%\includegraphics[height=\imgheight, keepaspectratio]{./figures/dnn/mei-matches/matched/layer4_1_conv2_262_86.png}
%&
%\includegraphics[height=\imgheight, keepaspectratio]{./figures/dnn/mei-matches/unmatched/layer4_1_conv2_1_284.png}\hspace{0.01\linewidth}%
%\includegraphics[height=\imgheight, keepaspectratio]{./figures/dnn/mei-matches/unmatched/layer4_1_conv2_0_116.png}
%\\

\end{tabular}

\captionof{figure}{\textbf{Visualization of (Un)matched Units Using Maximally Exciting Images.} We show MEIs for two layers of ResNet-$18$, comparing units classified as matched or unmatched by the partial soft-matching metric. Matched examples are sampled from the top $10\%$ of partial soft-matching scores; unmatched examples are sampled from the bottom $10\%$.}
\label{fig: mei-matching}

\end{center}

%\subsection{Maximally Exciting Images}
%Maximally Exciting Images (MEIs)—synthetic stimuli optimized to maximize individual unit responses—provide an interpretable visualization of what each neuron ``looks for'' in its input~\citep{erhan2009visualizing, pierzchlewicz2023energy, walker2019inception, bashivan2019neural}. We use MEIs to validate whether our unbalanced soft-matching metric correctly identifies matched vs. unmatched units across networks.
%\input{iclr2026/figures-code/mei-results}
%We synthesize MEIs\footnote{Details of MEI synthesis are provided in Appendix~\ref{sec: mei-synthesis}.} for unit pairs from two ResNet-18 models trained with different random seeds, sampling from neurons ranked as highly-matched (top $10\%$ of transport mass) versus poorly-matched (bottom $10\%$) by our metric. Fig.~\ref{fig: mei-matching} shows striking differences: highly-matched pairs produce nearly identical MEIs, revealing that these units have converged on similar feature detectors despite independent training. In contrast, unmatched pairs yield divergent MEIs with distinct visual patterns, confirming they likely implement different computations.
%
\subsection{Testing for privileged axes in aligned neural subpopulations}
%\begin{figure}[htbp!]
%    \centering
%    \includegraphics[width=\linewidth]{iclr2026/figures/dnn/privileged_axes_resnet18.png}
%    \caption{\textbf{Testing for Privileged Coordinate Systems Across Matched Neural Subpopulations.}  Alignment scores between ResNet-18 models under original and randomly rotated coordinate systems, across network layers and matched-mass thresholds. Rotation reduces alignment at all thresholds—including among the best-matched units—supporting convergence to a shared coordinate system even among the most aligned subpopulations. }
%    % \textbf{Testing for Privileged Coordinate Systems Across Matched Neural Subpopulations.} We examine whether coordinate system alignment persists when focusing on increasingly well-matched neural subsets between two ResNet-$18$ models trained from different random initializations. Each curve shows layer-wise alignment versus the fraction of included neurons; solid $=$ original basis, dashed $=$ after a random orthogonal rotation. The persistent gap between solid and dashed curves---including those for the best-matched neurons---indicates convergence to a similar, privileged coordinate system rather than arbitrary rotated bases, across layers and thresholds suggesting computational convergence at the single-unit level.
%    % }
%    \label{fig: privileged-axes}
%\end{figure}

%\begin{figure}[htbp!]
\begin{wrapfigure}{L}{0.75\linewidth}
    \centering
    \includegraphics[width=0.9\linewidth]{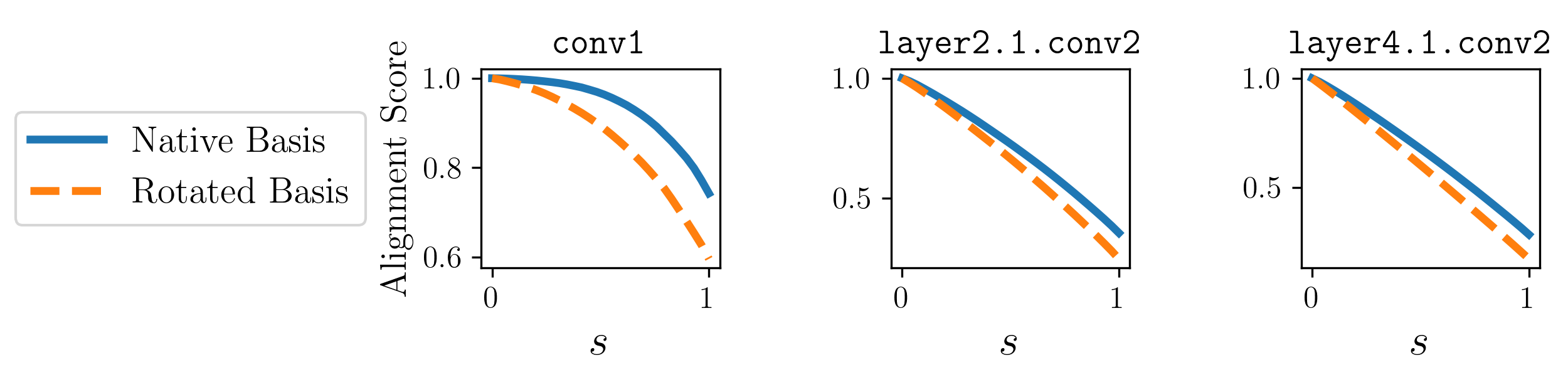}
    \caption{Alignment between ResNet-18 models under original and randomly rotated coordinate systems, across early, middle and later layers and matched-mass thresholds. Rotation reduces alignment at all thresholds---including among the best-matched units---supporting convergence to a shared coordinate system even among the most aligned subpopulations. }
    \label{fig: privileged-axes}
\end{wrapfigure}
%\end{figure}
Neural networks could in principle encode information in arbitrarily rotated coordinate systems, yet recent evidence suggests they converge on specific ``privileged'' axes. Networks trained from different initializations not only share representational geometry but actually align their coordinate systems---with individual neurons implementing similar computations across networks~\citep{khosla2024soft, khosla2024privileged, kapoor2025bridging}. This privileged basis hypothesis suggests that certain directions in activation space are preferred, potentially due to architectural constraints, \emph{e.g.,} axis-aligned nonlinearities (ReLU). We ask whether this alignment holds when restricted to the best-matched neurons. Using partial soft-matching, we partition increasingly well-matched neurons and test for privileged axes from the full population down to the strongest matched pairs.
%Neural networks could theoretically represent the same information using arbitrarily rotated coordinate systems---yet recent evidence suggests they instead converge on specific, ``privileged'' axes for encoding information. Prior work has shown that networks trained from different initializations not only learn similar representational geometries (measurable by rotation-invariant metrics like CKA), but actually align their coordinate systems—with individual neurons implementing similar computations across networks~\citep{khosla2024soft, khosla2024privileged, kapoor2025bridging}. This privileged basis hypothesis suggests that certain directions in activation space are systematically preferred, potentially due to architectural constraints like axis-aligned nonlinearities (ReLU) or optimization dynamics. We extend this investigation by asking: does coordinate system alignment persist even when we focus exclusively on the most well-matched neurons? Using unbalanced soft-matching, we can systematically partition neurons into increasingly aligned subsets and test whether privileged axes exist at every level of correspondence---from analyzing all neurons down to examining only the most strongly matched pairs.

For two ImageNet-trained ResNet-$18$ models initialized with different random seeds, we extract representations $\{\bm{X}_1, \bm{X}_2\}$ at each convolutional layer. To test for privileged axes, we apply a random orthogonal transformation $\bm{Q}$ (sampled uniformly from the Haar measure) to one network's representation and measure how this rotation affects alignment $s_{\mathrm{partial}}(\bm{X}_1\bm{Q}, \bm{X}_2)$. If neurons were arbitrarily oriented, this rotation would not affect the alignment. However, if a privileged basis exists, rotating away from it should decrease alignment.

We test multiple regularization values $(s)$ to sample subpopulations of varying alignment quality. Fig.~\ref{fig: privileged-axes} reveals that alignment consistently decreases under rotation across for all $s$ and depths, with an identical pattern across all convolutional layers (Appendix~\ref{sec: privileged-axes-all}). This demonstrates that privileged coordinate systems persist even among the most aligned neural subpopulations---the subset we might expect to be most robustly matched. The persistence of this effect suggests that coordinate alignment is not merely a statistical artifact of analyzing many neurons together, but reflects true convergence on similar computational solutions at the single-unit level.

%We perform this test at multiple regularization values ($m_{\mathrm{reg}}$), allowing us to examine subpopulations of varying alignment quality. Fig.~\ref{fig: privileged-axes} reveals that alignment consistently decreases under rotation across all regularization values and network depths. We demonstrate an identical pattern across all convolutional layers in Appendix~\ref{sec: privileged-axes-all}. This demonstrates that privileged coordinate systems persist even among the most aligned neural subpopulations---the subset we might expect to be most robustly matched. The persistence of this effect suggests that coordinate alignment is not merely a statistical artifact of analyzing many neurons together, but reflects true convergence on similar computational solutions at the single-unit level.

% discussion
\section{Discussion}
We introduced partial soft-matching, extending OT-based representational comparisons to account for partial correspondence between neural populations. This addresses a key limitation of methods that force all units into alignment, which can obscure genuine matches in the presence of noise. Simulations show that the method preserves true correspondences under noise and selects the correct model in system identification tasks. In fMRI data, it excludes low-reliability voxels and improves alignment precision across homologous brain regions. In deep networks, matched units exhibit shared MEIs, while unmatched units differ qualitatively. In both domains, partial soft-matching provides a more efficient way to order units by alignment quality, closely matching brute-force ablations, requiring only a single optimization at each chosen mass regularization value.

Some limitations remain. The L-curve heuristic for selecting matched mass performs well empirically, but its generality is unclear. We list some good practices that a practitioner should keep in mind while using the L-curve method in Appendix~\ref{appendix: best-practices}. Alternate strategies (\emph{e.g.,} area under the alignment-regularization curve)---may offer more robust summarization across multiple regularization values. Because partial OT relaxes mass conservation and violates the triangle inequality, it should be understood as a comparative tool rather than a strict metric. However, we note that recent theoretical work has developed partial Wasserstein variants that preserve full metric properties, including the triangle inequality~\citep{raghvendra2024new}. Future extensions could integrate these formulations for applications requiring strict metric axioms, such as clustering analyses. Although significantly faster than brute-force baselines, the $\mathcal{O}(n^3\log{n})$ cost can limit scalability to very large datasets. These considerations aside, this work highlights that meaningful comparison does not require complete unit overlap: partial soft-matching enables principled analysis of convergent and divergent representational structure across neural systems.
%Some limitations remain. The L-curve heuristic for selecting matched mass performs well empirically, but its generality is unclear. Alternate strategies---such as computing the area under the alignment-versus-regularization curve—may offer more robust summarization across multiple regularization values. Because partial OT relaxes mass conservation and violates the triangle inequality, it should be understood as a comparative tool rather than a strict metric. And while far more efficient than brute-force baselines, the $\mathcal{O}(n^3 \log n)$ cost may still limit scalability for very large networks. These considerations aside, this work highlights that meaningful comparison does not require complete unit overlap: unbalanced soft-matching enables principled analysis of both convergent and divergent representational structure across neural systems.

\bibliography{iclr2026_conference}
\bibliographystyle{iclr2026_conference}

\appendix
\renewcommand{\thesection}{A\arabic{section}}
\renewcommand{\thefigure}{A\arabic{figure}}
\renewcommand{\thetable}{A\arabic{table}}
\setcounter{figure}{0}  
\setcounter{table}{0}
\setcounter{section}{0}
\newpage
\section{Appendix}
\subsection{Best Practices}
\label{appendix: best-practices}
We treat the L-curve elbow as a practical heuristic for selecting the matched mass $s$, rather than a formal rule. As with any hyperparameter in machine learning, the L-curve should be treated as a user-dependent choice.

The L-curve typically fails when the cost-regularization curve $\left(\zeta(s), \rho(s)\right)$ is smooth and monotonic. In this regime, the estimated curvature is uniformly small, and any computed \emph{``inflection''} is likely an artifact of either numerical noise or local smoothness. A common empirical signature of this failure mode is that the algorithm selects an ``optimal'' regularization $s$ at either of the tail ends of the $\left(\zeta, \rho\right)$ curve. When this occurs, we suggest a simple diagnostic---one can visually inspect the L-curve and check the magnitude of curvature at the selected point. However, if the curvature profile is non-informative, one can use alternative rank summary statistics such as area under the $\left(\zeta, \rho\right)$ curve.
\subsection{Comparison of Neural Recordings for an Alternate Subject Pair}
\label{sec: noise-ceiling-nsd-alt}
\begin{figure}[htbp!]
    \centering
    \includegraphics[width=\columnwidth]{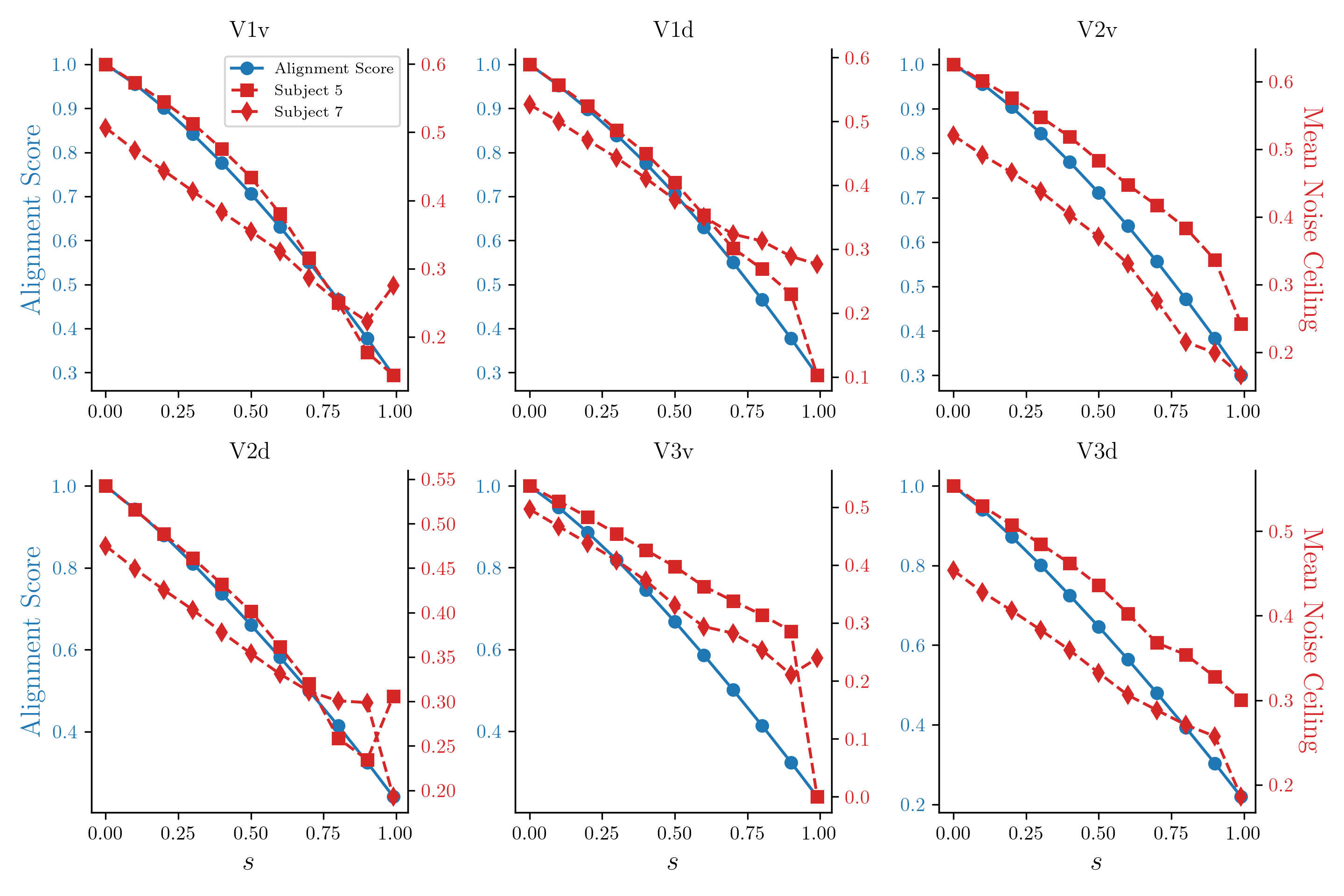}
    \caption{\textbf{Aligning Voxel Responses Between Subject IDs 5 and 7 in NSD.} We plot the partial soft-matching alignment score achieved at different mass regularization values and the mean noise ceilings of the voxels that were kept at each regularization. For The alignment criterion consistently identifies low noise-ceiling voxels for exclusion for Subjects $5$ and $7$.}
    \label{fig:nsd-nc-subj5-7}
\end{figure}

\subsection{Generation of Synthetic Data}
\label{sec: synth-generation}
For the synthetic experiments in Section~\ref{sec: synthetic}, we construct two approximately one-to-one matched ``neural representations'' by linearly mixing a small subset of latent factors with sparse coefficients. We first generate a factor matrix $\bm{F}\in\mathbb{R}^{m\times k}$ whose columns denote the responses of $k$ linearly independent factors across $m$ unique stimuli. Each column is drawn i.i.d from $\mathcal{N}(0, 1)$, and the matrix is orthogonalized using the Gram-Schmidt procedure. We then generate a pair of representations $\bm{Z}_1\in\mathbb{R}^{m\times n}$ and $\bm{Z}_2\in\mathbb{R}^{m\times n}$ as sparse linear mixtures of these factors:
\begin{align*}
    \bm{X}_1 &= [\bm{X}_1]_{ij}\sim\mathcal{N}(0,1),\;\;
    \bm{X}_2 = [\bm{X}_2]_{ij}\sim\mathcal{N}(0,1)\\
    \bm{M}_1 &= \bm{S}\odot\bm{X}_1, \;\; \bm{M}_2 = \bm{S}\odot\bm{X}_2 \;\; \textrm{where} \;\; \bm{S}\in\{0, 1\}^{k\times n}\\
    \bm{Z}_1 &= \bm{FM}_1, \;\; \bm{Z}_2 = \bm{FM}_2
\end{align*}
where $\odot$ denotes the Hadamard (element-wise) product. The binary mask $\bm{S}$ controls the degree of sparsity. By using the \emph{same} support mask for both populations, each column of $\bm{Z}_1$ and the corresponding column of $\bm{Z}_2$ depend on the same subset of factors but with independent mixing weights. Choosing $\bm{S}$ to be highly sparse makes each ``neuron'' depend on only one (or a combination) of factors, thereby producing clear, approximately one-to-one correspondences across the two populations with the intent of mimicking selective tuning to a small set of features.

To introduce outliers and measurement noise, we augment each population with additional \emph{``noise neurons''}. Concretely, we append $n'_1$ and $n'_2$ columns of i.i.d Gaussian noise $\varepsilon\sim\mathcal{N}(0, 1)$, yielding $\bm{Z}_1\in\mathbb{R}^{m\times (n+n'_1)}$ and $\bm{Z}_2\in\mathbb{R}^{m\times(n+n'_2)}$. 

\subsection{Baseline Matching Algorithms}
\label{sec: baseline-matching-algos}
In this section, we describe the baseline matching algorithms used to rank-order the tuning similarities as demonstrated in Sec.~\ref{sec: baseline-comparison}. In all cases, we consider a representation pair $\bm{Z}_1\in\mathbb{R}^{m\times n_1}$ and $\bm{Z}_2\in\mathbb{R}^{m\times n_2}$ where $m$ is the number of stimuli and $\{n_1, n_2\}$ are the number of neurons respectively.

\paragraph{Brute-Force Matching.} We construct a greedy baseline for ordering the deletion of neurons based on their soft-matching correlation score. Starting from a complete set of $N$ neurons, we establish a baseline score. At each iteration, we evaluate, for every remaining neuron $i$, the score obtained after removing neuron $i$ and re-fitting the soft-match transform on the reduced representation. We then remove the neuron whose deletion produces the largest decrease in the matching distance (equivalently, the largest improvement in the alignment score if removal improves the score), append it to the deletion order, and repeat on the remaining neurons. Thus, in essence, we construct a \emph{rank-ordering} of neurons in terms of their tuning similarities based on the soft-matching objective. 
\begin{algorithm}
\caption{Brute-Force Matching}
\begin{algorithmic}[1]
\State $R \leftarrow \{1,\dots,N\}$ \Comment{set of remaining neuron indices}
\State $\overline{s}\leftarrow \mathsf{SoftMatch}(\bm{Z}_{1_:,R},\,\bm{Z}_{2:,R})$ \Comment{baseline score on full set}
\State $\pi \leftarrow [\,]$ \Comment{initialize deletion ordering}
\While{$R\neq \varnothing$}
  \For{each $i\in R$}
    \State $R_{-i}\leftarrow R\setminus\{i\}$
    \State $s_i \leftarrow \mathsf{SoftMatch}\bigl(\bm{Z}_{1_:,R_{-i}},\,\bm{Z}_{2:,R_{-i}}\bigr)$ \Comment{re-fit soft matching without neuron $i$}
    \State $\Delta_i \leftarrow s_i - \overline{s}$ \Comment{change in score produced by deleting $i$}
  \EndFor
  \State $i^\star \leftarrow \arg\min_{i\in R} \Delta_i$ \Comment{pick neuron whose deletion most decreases score}
  \State append $i^\star$ to the end of list $\pi$
  \State $R \leftarrow R\setminus\{i^\star\}$ \Comment{permanently remove neuron}
  \State $\overline{s}\leftarrow s_{i^\star}$ \Comment{update current score to the one after deletion}
\EndWhile
\State \Return $\pi$ \Comment{deletion order from least $\rightarrow$ most matched}
\end{algorithmic}
\end{algorithm}

\paragraph{Correlation-Based Matching.} For each fitted soft-matching plan $\bm{T}$ on a response pair, we perform a bidirectional correlation-based voxel matching. We project responses from one representational pair into the others space $\widetilde{\bm{Z}_1} = \bm{Z}_1\bm{T}$ and compute the Pearson correlation between the response pair $\texttt{corr}(\widetilde{\bm{Z}_1}, \bm{Z}_2)$. We retain the top-$k$ correlated units in $\bm{Z}_2$, where $k$ is determined by the number of units (un)matched using the partial soft-matching distance to maintain consistency during comparison. We repeat this procedure in the reverse direction $\widetilde{\bm{Z}_2} = \bm{Z}_2\bm{T}^\top$ and compute Pearson correlations $\texttt{corr}(\bm{Z}_1, \widetilde{\bm{Z}}_2)$ to find the matched units.

\begin{algorithm}
\caption{Correlation-Based Matching}
\label{alg:voxel_selection}
\begin{algorithmic}[1]
    \Statex \textbf{Forward mapping (response 1 $\to$ response 2):}
    \State $\widetilde{\bm{Z}_1} \gets \bm{Z}_1\bm{T}$
    \State $c_{1\rightarrow 2} \gets \texttt{corr}\left(\widetilde{\bm{Z}}_1, \bm{Z}_2\right)$
    \State $\textrm{kept}_2 \leftarrow \texttt{argsort}(c_{1\rightarrow2})[:k]$
    
    \Statex \textbf{Reverse mapping (response 2 $\to$ response 1):}
    \State $\widetilde{\bm{Z}_2} \gets \bm{Z}_2\bm{T}^\top$
    \State $c_{2\rightarrow 1} \gets \texttt{corr}\left(\bm{Z}_1, \widetilde{\bm{Z}}_2\right)$
    \State $\textrm{kept}_1 \leftarrow \texttt{argsort}(c_{2\rightarrow1})[:k]$
\end{algorithmic}
\end{algorithm}

\paragraph{Partial Soft-Matching.} For a given regularization value $s$, we fit a partial soft-matching transport plan $\bm{T}\in\mathbb{R}^{n_1\times n_2}$ between a response pair $\{\bm{Z}_1, \bm{Z}_2\}$. We compute the outgoing mass $r_i = \sum_i\bm{T}_{ij}$ for each source unit and incoming mass $c_j = \sum_j\bm{T}_{ij}$ for each target unit, and retain only those units whose total mass exceeds a small threshold $\tau$, serving as a way to determine the unmatched units. We repeat this procedure over a grid of regularization values $\mathcal{S}\leftarrow\{s_1,\cdots,s_k\}$, yielding ``per-$s$'' sets of \emph{kept} units $\mathcal{K}_1(s)$ and $\mathcal{K}_2(s)$ that are matched.

\begin{algorithm}
\caption{Partial Soft-Matching}
\label{alg:partial-matching}
\begin{algorithmic}[1]
\State $\mathcal{S} \leftarrow \{s_1,\cdots, s_k\}$ \Comment{initialize a list of regularization values}
\State $\tau\leftarrow \texttt{1e-6}$ \Comment{initialize a threshold for transport weight}
\For{$m \in \mathcal M$}
    \State $\bm{T} \leftarrow \mathsf{ParSM}(m_{\mathrm{reg}}=s)$  \Comment{compute optimal transport plan} 
    \State  $\;r_i \leftarrow \sum_{j=1}^{n_2} \bm{T}_{ij}$ for $i=1,\dots,n_1$ \Comment{outgoing mass per source unit}
    \State  $\;c_j \leftarrow \sum_{i=1}^{n_1} \bm{T}_{ij}$ for $j=1,\dots,n_2$ \Comment{incoming mass per target unit}
    \State $\mathcal K_1 \leftarrow \{\, i \;|\; r_i \ge \tau\,\}$
    \State $\mathcal K_2 \leftarrow \{\, j \;|\; c_j \ge \tau\,\}$
    \State \Return $\mathcal K_1,\mathcal K_2$ 
\EndFor
\end{algorithmic}
\end{algorithm}

\subsection{Synthesis of Maximally Exciting Images}
\label{sec: mei-synthesis}
Given a CNN $f\colon \bm{\mathcal{X}}\to\mathbb{R}^K$ that maps an input image $\bm{x}\in\bm{\mathcal{X}}\subset\mathbb{R}^{h\times w\times c}$ to $K$ class logits, we define a scalar target $g(\bm{x})$ as the activation of the unit of interest (\emph{e.g.,} feature-map channel, readout). For convolutional layers, when aligning representations, we evaluate channels at the \emph{center} spatial location, motivated by evidence that convolutional feature maps are equivalent up to a circular shift~\citep{williams2021generalized, kapoor2025bridging}. Fixing the center neuron across channels thus allows us to consistently describe representations. We synthesize one image per channel by solving:
\begin{equation*}
    \bm{x}^\star = \argmax_{\bm{x}\in\bm{\mathcal{X}}}\left(\mathbb{E}_{\tau\sim\mathcal{T}}g(\tau(\bm{x}))- \sum_r \lambda_r R_r(\bm{x})\right)
\end{equation*}
where $\mathcal{T}$ is a distribution over input transformations (\emph{e.g.,} jitter, crop). Each $R_r(\cdot)$ is a regularizer with weight $\lambda_r$. In practice, we sample a new $\tau$ at each iteration and maximize the objective via gradient ascent at the center pixel of every channel. We implement this optimization with the Lucent library \footnote{\href{https://github.com/TomFrederik/lucent/}{https://github.com/TomFrederik/lucent/}} using total variation (TV) as the regularizer. Each synthesized image is of shape $224\times 224\times 3$.

\newpage
\subsection{Privileged Axes Persists in All Kernel Subpopulations}
\label{sec: privileged-axes-all}
\begin{figure}[htbp!]
    \centering
    \includegraphics[width=\linewidth]{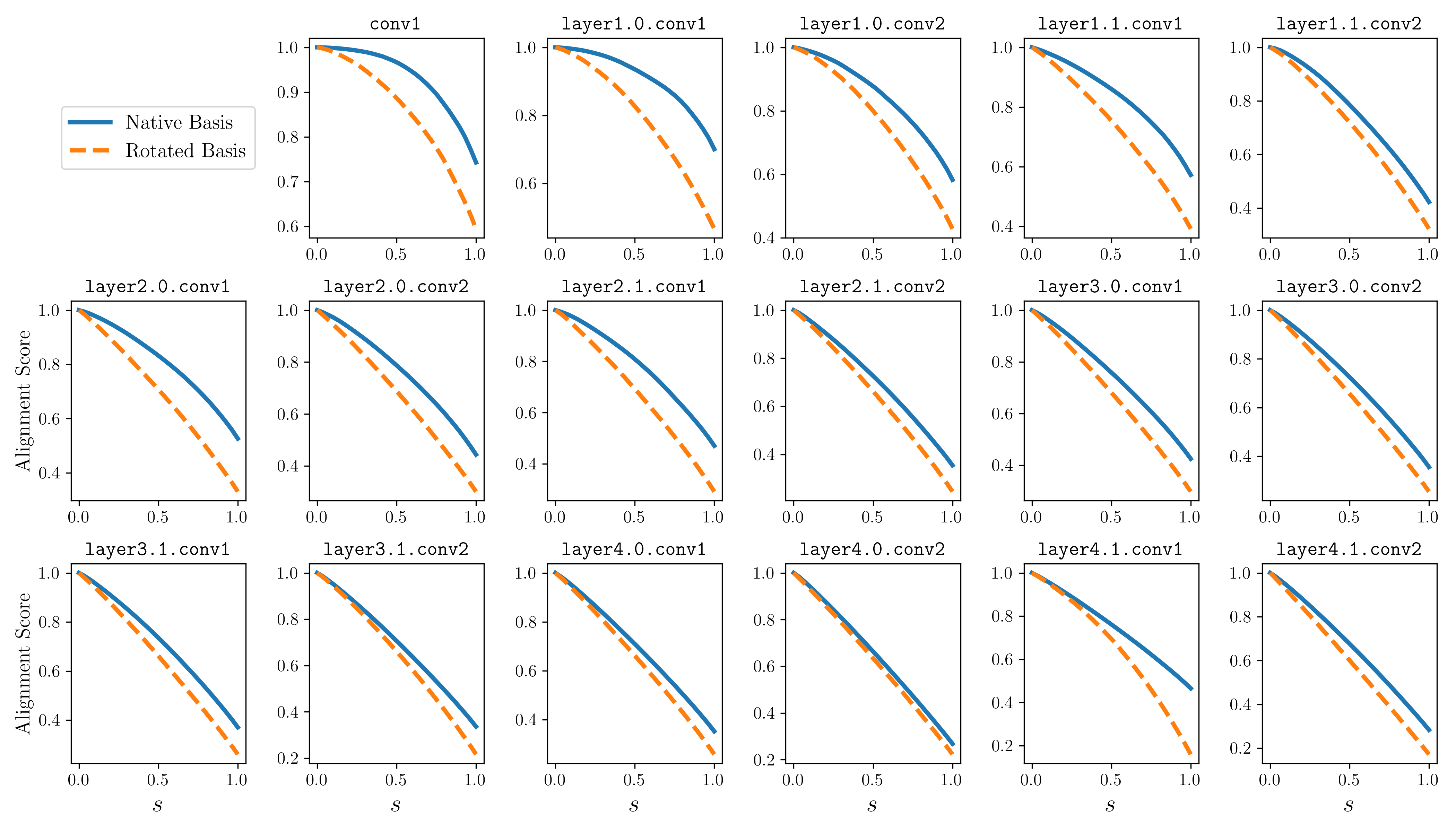}
    \caption{\textbf{Testing for Privileged Coordinate Systems Across Matched Neural Subpopulations.} For all convolutional layers in a pair of ImageNet-trained ResNet-$18$ models, we show that a privileged solution basis persists.}
    \label{fig:privileged-axes-all}
\end{figure}

\subsection{Additional Results For Matched Maximally Exciting Images}
\label{sec: mei-extra}
In the following section, we show additional matched MEI pairs for two layers in a ResNet-$18$, while still displaying the top $10\%$ and bottom $10\%$ examples in Fig.~\ref{fig: mei-matching-appendix}.
%\newlength{\imgheight}
\setlength{\imgheight}{1.75cm}
\begin{center}
\begin{tabular}{@{} p{0.47\textwidth} | p{0.47\textwidth} @{}}

% column headers
\multicolumn{1}{@{}p{0.47\textwidth}@{}}{\centering{\textbf{\fontfamily{cmr}\selectfont Matched}}} &
\multicolumn{1}{@{}p{0.47\textwidth}@{}}{\centering{\textbf{\fontfamily{cmr}\selectfont Unmatched}}} \\[4pt]

% row 1 
\centering
% left column
\includegraphics[height=\imgheight, keepaspectratio]{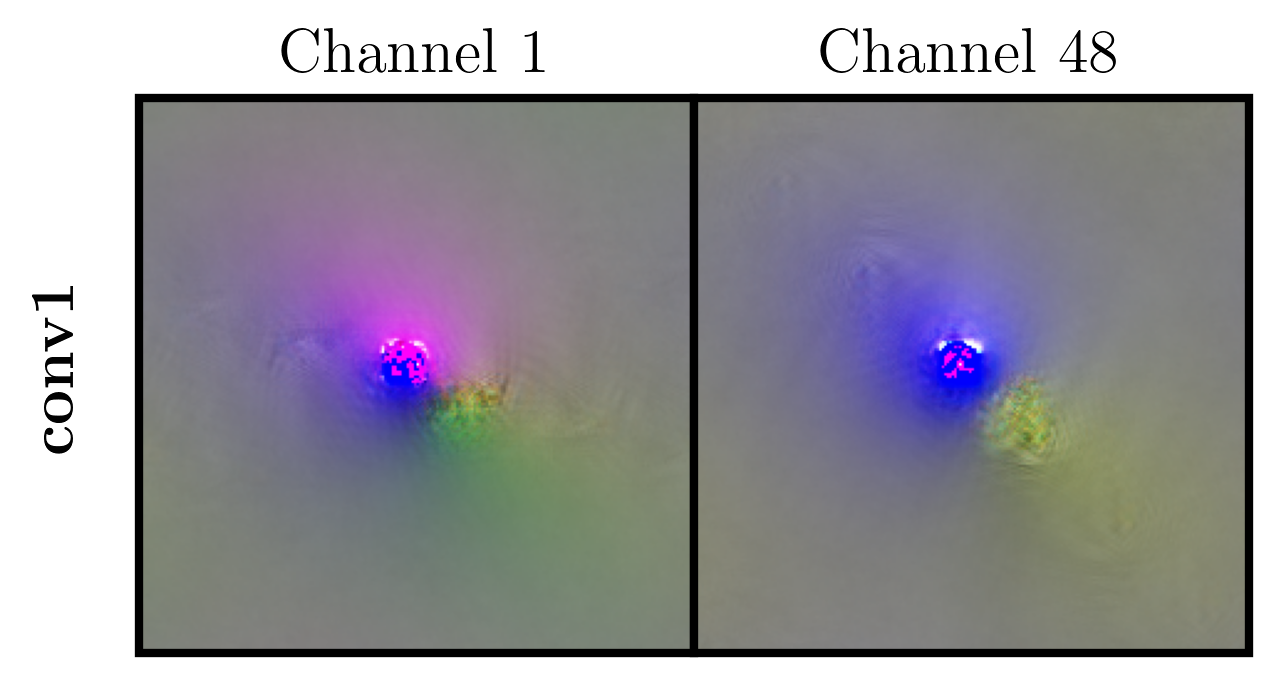}\hspace{0.01\linewidth} 
\includegraphics[height=\imgheight, keepaspectratio]{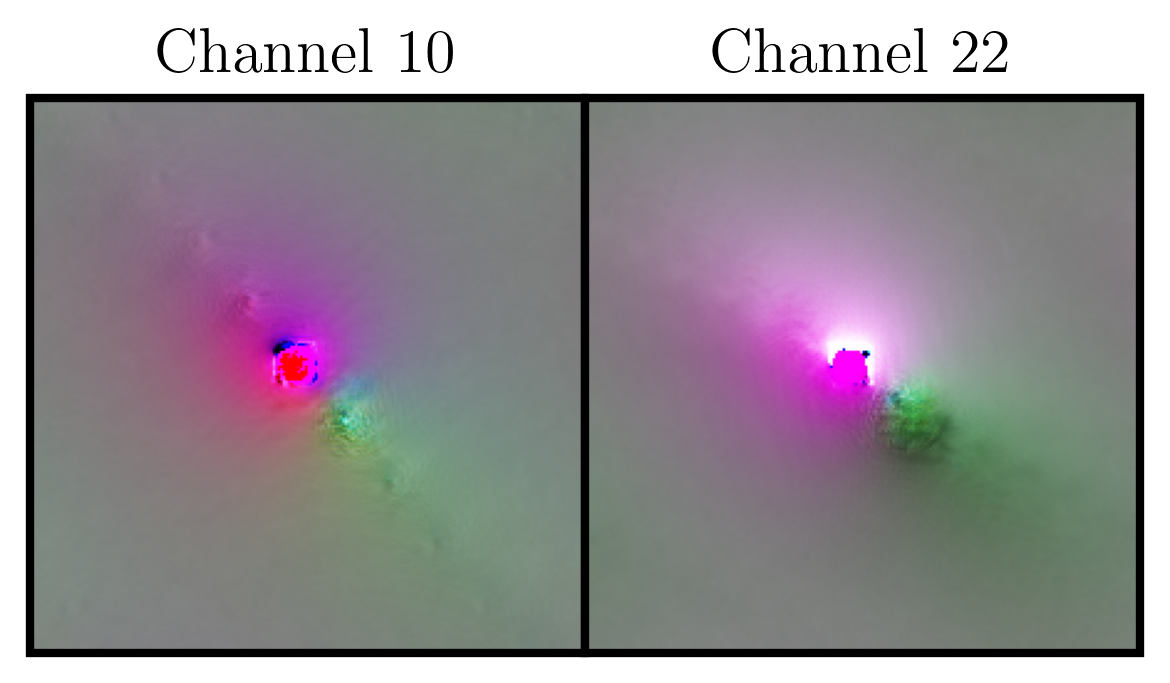}
&
% right outer column
\includegraphics[height=\imgheight, keepaspectratio]{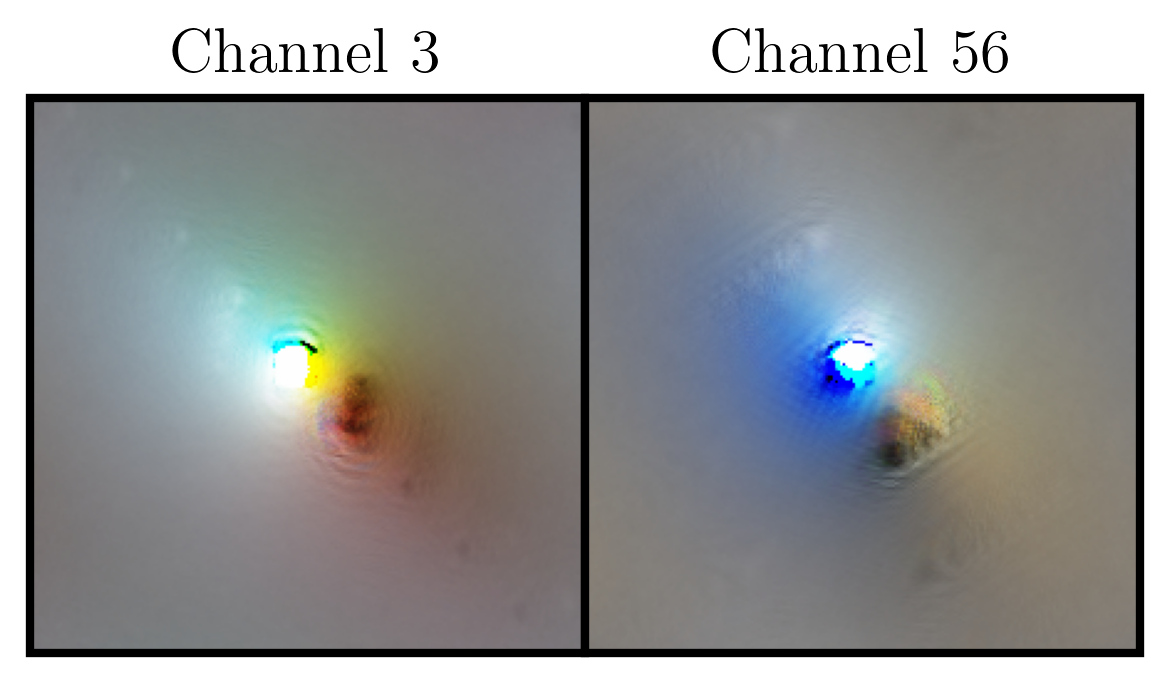}\hspace{0.01\linewidth}%
\includegraphics[height=\imgheight, keepaspectratio]{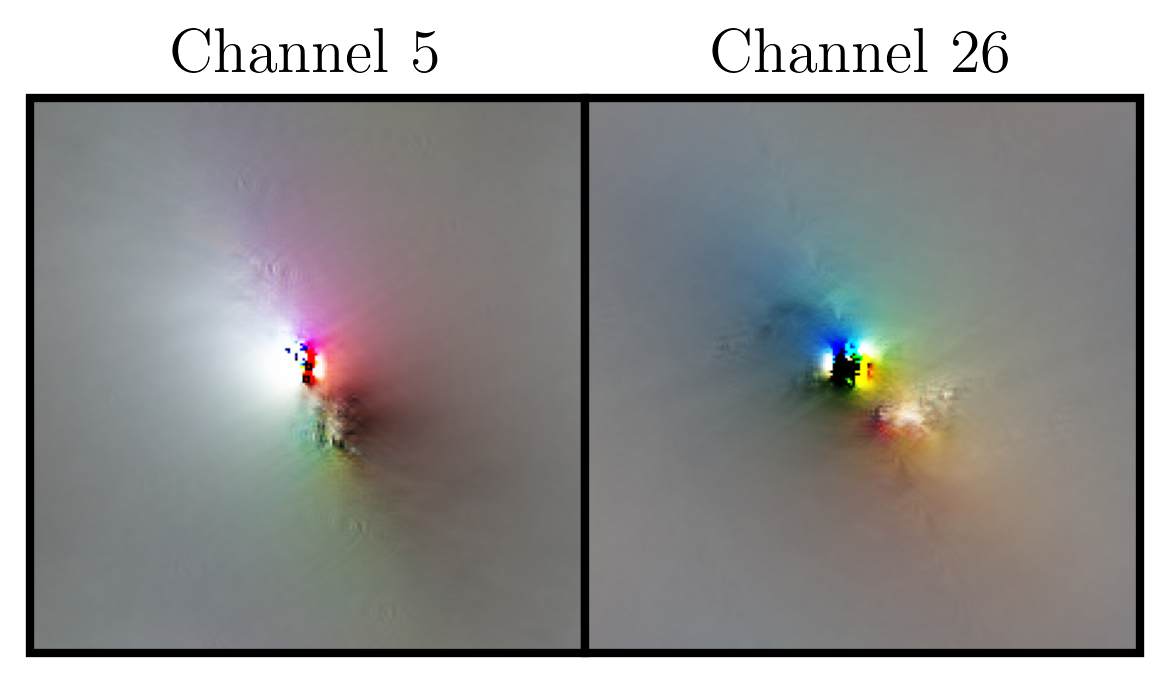}
\\[1pt]

% row 2 
\centering
\includegraphics[height=\imgheight, keepaspectratio]{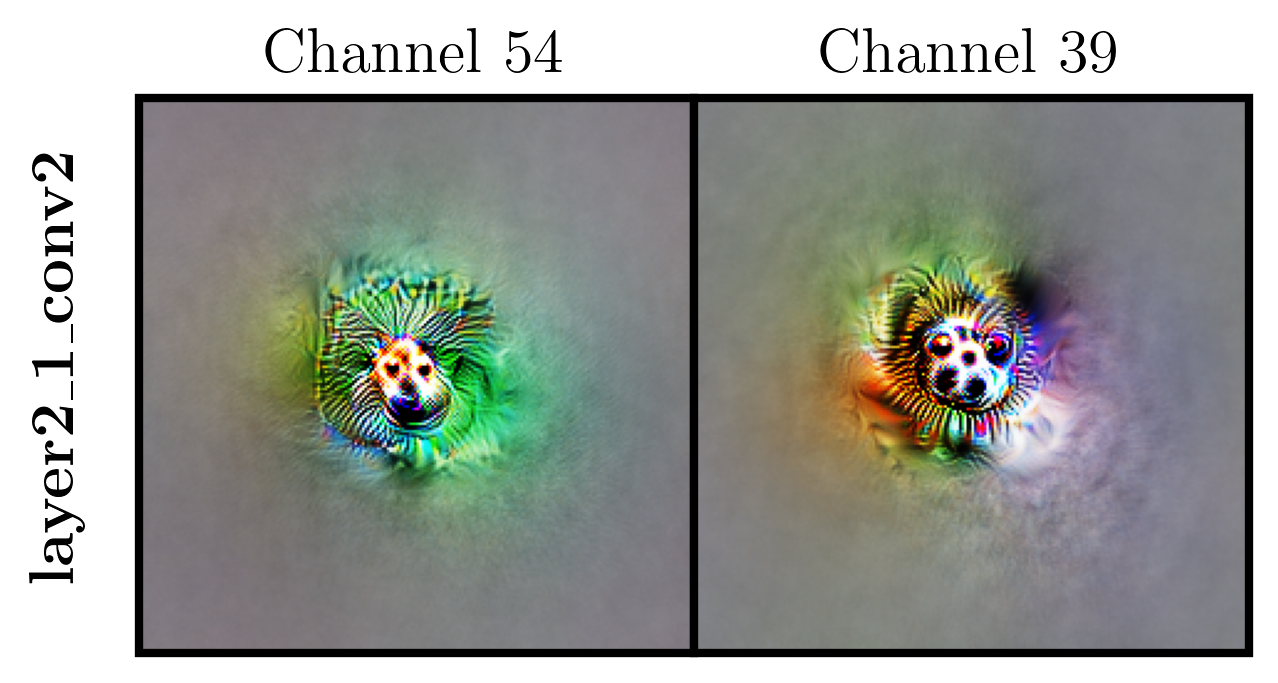}\hspace{0.01\linewidth}%
\includegraphics[height=\imgheight, keepaspectratio]{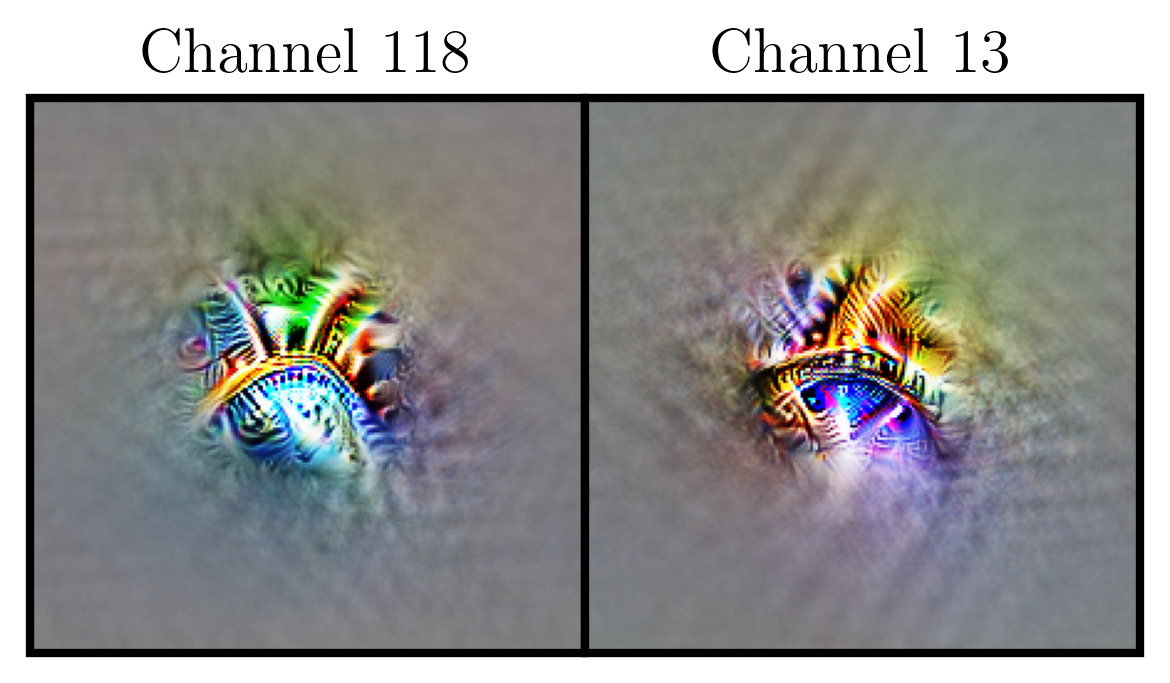}
&
\includegraphics[height=\imgheight, keepaspectratio]{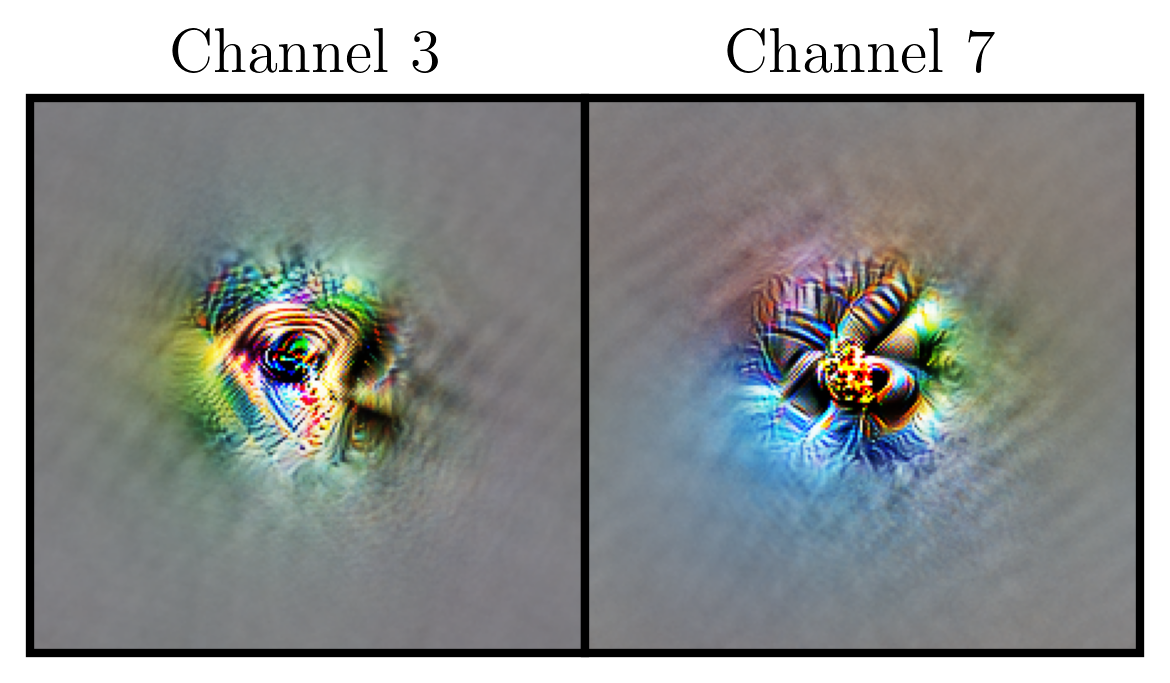}\hspace{0.01\linewidth}%
\includegraphics[height=\imgheight, keepaspectratio]{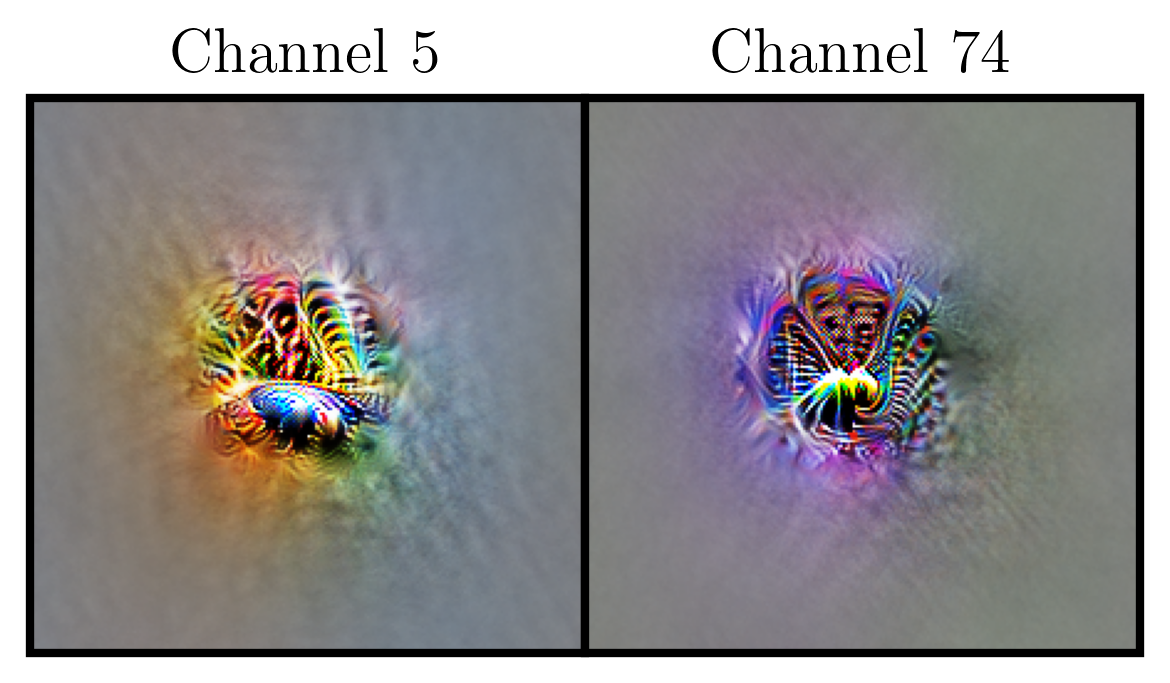}
\\%[1pt]

\end{tabular}

\captionof{figure}{\textbf{Additional visualizations of (Un)matched Units Using Maximally Exciting Images.} We show additional MEIs for two layers of ResNet-$18$, comparing more pairs of top matched and unmatched units using the partial soft-matching metric.}
\label{fig: mei-matching-appendix}

\end{center}

\newpage
\subsection{Sensitivity To Choice of Cost Function}
\label{appendix: euclidean}
\begin{figure}[htbp!]
    \centering
    \includegraphics[width=\linewidth]{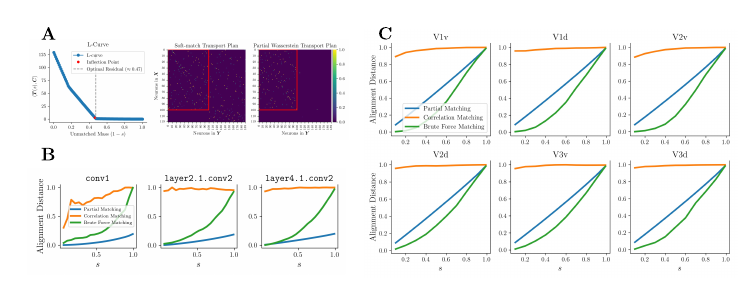}
    \caption{\textbf{Partial Soft-Matching Using a Euclidean Cost Function} \textbf{(A)} We visualize the L-curve elbow computed using a Euclidean cost function (\textbf{left}), and the corresponding transport plans (\textbf{right}). We find that the optimal regularization value $s$ and transport plans are identical to that computed using a cosine similarity cost function. \textbf{(B)}  Rank-ordering of neurons from $3$ layers (early, middle and late) in an ImageNet-trained ResNet-$18$ network using a squared Euclidean cost function. Removing units by using the Euclidean distance yields identical trends as using a cosine similarity cost. \textbf{(C)} We rank-order voxels from a subject pair in NSD using six visual areas by using a squared Euclidean cost function. Ordering voxels using a Euclidean distance cost function also reveals identical trends as using cosine similarity.}
    \label{fig: all-euclidean}
\end{figure}

For all results demonstrated in Sec.~\ref{sec: baseline-comparison}, we use cosine similarity to rank-order units. In this section, we construct a squared-Euclidean distance cost matrix (i.e.: $\bm{C}_{ij} = ||\bm{x}_i - \bm{y}_j||^2$) to compute the optimal transport plan. In synthetic experiments (Fig.~\ref{fig: all-euclidean}-A), deep neural networks (Fig.~\ref{fig: all-euclidean}-B) and brain data (Fig.~\ref{fig: all-euclidean}-C), we find our conclusions remain unaffected by the choice of cost function. For the brain data and DNN alignment plots, we normalize the distances by their maximum value such that values can be visualized in the same plot.

%\begin{figure}[htbp!]
%    \centering
%    \includegraphics[width=1.\linewidth]{iclr2026/figures/nsd-matching-baselines-euclidean.png}
%    \caption{\textbf{Rank-Order Voxels in Brain Data Using Euclidean Distance.} \textcolor{red}{We rank-order voxels from a subject pair in NSD using six visual areas by using a squared Euclidean cost function. Ordering voxels using a Euclidean distance cost function reveals identical trends as using cosine similarity.}}
%    \label{fig: euclidean-nsd}
%\end{figure}
%
%\begin{figure}[htbp!]
%    \centering
%    \includegraphics[width=1.\linewidth]{iclr2026/figures/resnet18-matching-baselines-in1k-euclidean.png}
%    \caption{\textbf{Rank-Order Neurons in ResNet18 Using Euclidean Distance.} \textcolor{red}{We rank-order neurons from $3$ layers (early, middle and late) in an ImageNet-trained ResNet-$18$ network using a squared Euclidean cost function. Removing units by using the Euclidean distance yields identical results as using a cosine similarity cost.}}
%    \label{fig:placeholder}
%\end{figure}

\end{document}